\documentclass[reqno,10pt]{amsart}
\usepackage{graphics}
\usepackage{amsmath,amssymb,amsfonts}
\usepackage{mathrsfs}
\usepackage{wasysym}
\usepackage{algorithm}
\usepackage{algorithmic}

\usepackage{multirow,booktabs,comment}
\usepackage{epstopdf}

\usepackage{todonotes}
\usepackage[normalem]{ulem}

\usepackage{xcolor}
\usepackage{soul}

\usepackage[colorlinks=true,urlcolor=blue,citecolor=red,
anchorcolor=black,linkcolor=blue]{hyperref}

\usepackage{subcaption}

\usepackage{lineno}

\newtheorem{theorem}{Theorem}
\theoremstyle{plain}

\newtheorem{remark}[theorem]{Remark}

\numberwithin{equation}{section}
\numberwithin{equation}{section}
\numberwithin{theorem}{section}

\newcommand{\Bx}{\boldsymbol{x}}
\newcommand{\By}{\boldsymbol{y}}

\newcommand{\BY}{\boldsymbol{Y}}
\newcommand{\BZ}{\boldsymbol{Z}}
\newcommand{\Bs}{\boldsymbol{s}}
\newcommand{\Bt}{\boldsymbol{t}}
\newcommand{\Ba}{\boldsymbol{a}}
\newcommand{\Bb}{\boldsymbol{b}}
\newcommand{\Bz}{\boldsymbol{z}}

\newcommand{\Btheta}{\boldsymbol{\theta}}
\newcommand{\BI}{\mathbf{I}}

\newcommand{\xw}[1]{{\color{blue}{#1}}}

\begin{document}
	
\title[]{Coupling the reduced-order model and the generative model for an importance sampling estimator}
\author{Xiaoliang Wan and Shuangqing Wei} 
\curraddr[X. Wan]{Department of Mathematics, 
and Center for Computation and Technology, 
Louisiana State University\\
Baton Rouge, LA 70803}
\email[X. Wan]{xlwan@lsu.edu}
\curraddr[S. Wei]{Division of Electrical \& Computer Engineering, 
Louisiana State University\\
Baton Rouge, LA 70803}
\email[S. Wei]{swei@lsu.edu}

\begin{abstract}
In this work, we develop an importance sampling estimator by coupling the reduced-order model and the generative model in a problem setting of uncertainty quantification. The target is to estimate the probability that the quantity of interest (QoI) in a complex system is beyond a given threshold. To avoid the prohibitive cost of sampling a large scale system, the reduced-order model is usually considered for a trade-off between efficiency and accuracy. However, the Monte Carlo estimator given by the reduced-order model is biased due to the error from dimension reduction. To correct the bias, we still need to sample the fine model. An effective technique to reduce the variance reduction is importance sampling, where we employ the generative model to estimate the distribution of the data from the reduced-order model and use it for the change of measure in the importance sampling estimator. To compensate the approximation errors of the reduced-order model, more data that induce a slightly smaller QoI than the threshold need to be included into the training set. Although the amount of these data can be controlled by a posterior error estimate, redundant data, which may  outnumber the effective data, will be kept due to the epistemic uncertainty. To deal with this issue, we introduce a weighted empirical distribution to process the data from the reduced-order model. The generative model is then trained by minimizing the cross entropy between it and the weighted empirical distribution. We also introduce a penalty term into the objective function to deal with the overfitting for more robustness. Numerical results are presented to demonstrate the effectiveness of the proposed methodology. 

\end{abstract}

\maketitle
\section{Introduction}
Generative modeling has become a central object in modern machine learning. The goal of generative modeling is to model all dependencies within high-dimensional data using a full joint probability density function (PDF), and to generate new samples from the learned distribution. The ability to manipulate the joint PDF enables the probabilistic unsupervised learning of realistic world models. Generative modeling has found a wide range of applications such as image processing, speech synthesis, text analysis, etc. Significant advances have been achieved in the recent development of generative modeling. Typical approaches include variational autoencoders \cite{Kingma_2016}, autoregressive models \cite{Graves_2013,Oord_2016a,Oord_2016b,Papamakarios_2018}, flow-based generative models \cite{Dinh_2014,Dinh_2016,Dhariwal_2018}, and generative adversarial networks (GANs) \cite{Goodfellow_2014}. 

Meanwhile we note that multivariate density estimation is a classical topic in statistics \cite{Scott2015}, where it shows the equivalent sample size with respect to a dimensionless measure of accuracy will increase at least exponentially with respect to the dimensionality. In contrast to the thousands of dimensions considered in generative modeling, practical applications of  nonparametric density estimators in more than three dimensions often suffer a great deal from the curse of dimensionality. Although it is not quite fair to consider a direct comparison between nonparametric density estimators and generative models, where nonparametric density estimators focus on the asymptotic behavior of mean integrated square error while the generative models focus on the learning ability and flexibility, generative models, which can be regarded as parametric models, seems able to provide a very general representation of data like the nonparametric estimator, thanks to the capability of deep neural networks for high-dimensional nonlinear approximation. In this work, we are trying to understand if we are able to adapt the generative modeling into a problem setting of uncertainty quantification as a flexible means to establish communications between two mathematical models through data. In particular, we consider an importance sampling estimator
\[
\mathbb{E}_\rho[I_B]=\int \frac{I_B(\By)\rho(\By)}{\eta(\By)}\eta(\By)d\By=\mathbb{E}_{\eta}[I_B\frac{\rho}{\eta}],
\]  
where $\rho$ and $\eta$ are two PDFs, and $I_B$ is an indicator function in terms of the set $B$. {Each sample $\By$ may be related to the solution $u(t,\Bx,\BY)$ of a PDE subject to random inputs $\BY\in\mathbb{R}^n$, where $t$ and $\Bx$ indicate the time and space variable respectively. The random event $B$ is defined by a functional of $u(t,\Bx,\BY)$, e.g., the $L_2$ norm on a space-time domain is larger than a prescribed threshold. $\rho(\By)$ is the PDF of $\BY$ and $\eta(\By)$ is the candidate for the change of measure.} We assume that our best a prior knowledge of $\eta$ is given by a set of data such that we need to estimate the data distribution first before implementing the importance sampling estimator.  We will use a generative model to represent $\eta$. The study of generative modeling usually focuses on the minimization of a certain measure on the distance between the model and the data distribution while our main concern is the effectiveness of the importance sampling estimator which can be measured quantitatively by the degree of variance reduction. Due to the overfitting, $\eta$ that is closer to the data distribution might not introduce variance reduction.  Thus the robustness is an {important} issue in addition to the dimensionality of $\By$. Our problem setting requires an explicit evaluation of the density function, which makes the adaption of some generative models such as GAN and variational autoencode not straightforward. In this work, we will employ the flow-based generative models \cite{Dinh_2016,Dhariwal_2018}, which provide tractable likelihood and exact inference due to the invertible network. 

We will construct an importance sampling estimator using multi-fidelity models: one fine model and its reduced-order model. The goal is to obtain the probability $\Pr(B)$ or $\mathbb{E}[I_B]$ with respect to the fine model. However, since each sample corresponds to solving a large scale problem, which is time consuming, we want to collect some data from a reduced-order model, and use them to construct $\eta(\By)$ for the importance sampling on the fine model. To make the strategy practical, we have considered the following two issues: First, with respect to the fine model, there exist noise in the data from the reduced-order model, which means we cannot simply keep the data satisfying $B$ for the reduced-order model. We need to enlarge the data set to tolerate the error from model reduction. Unfortunately because of the epistemic uncertainty in the error of reduced-order model, redundant data might be kept. We {have} proposed a weighted empirical distribution such that the important data have a larger weight while the less important data have a smaller weight. We {then} approximate the weighted empirical distribution using a flow-based generative model. Second, {the importance sampling estimator may fail due to the overfitting in the training process of the generative model.}  This mainly an issue about regularization. When the data set is not large enough, extra regularization is needed other than that provided by the stochastic optimization. We will show that incorporating the properties of the problem can provide a much more robust regularization than the general regularization techniques such as early {stopping}. More specifically, we add a penalty term to balance the fact that the ratio $\frac{I_B\rho}{\eta}$ should be close to a constant for variance reduction and the minimization of the cross entropy. Because the flow-based generative model has an explicit density function, such a penalty term can be easily implemented. 

The paper is organized as follows. In the next section we specify the problem setting and develop a guiding principle for our methodology. In section \ref{sec:generative_model} we build up the flow-based generative model used in this work. The main numerical strategy is developed in section \ref{sec:wed}. Some details related to implementation are given in section \ref{sec:implementation}. We present numerical experiments in section \ref{sec:num} followed by a summary section. 

\section{Problem description}
We are interested in simulating the random events given by a partial differential equation (PDE) subject to uncertainty. We present our methodology using the following general mathematical model:
\begin{equation}\label{eqn:general_model}
\mathcal{L}(u(t,\Bx);\boldsymbol{\BY})=0,
\end{equation}
where $\mathcal{L}$ is a space-time differentiation operator, $t$ the time, $\Bx\in\mathbb{R}^d$ the space variable, and $\BY\in\mathbb{R}^n$ a $n$-dimensional random vector. Let $B=\{\By|g(u)>0\}$, where $g(\cdot)$ is a functional indicating the Quantity of Interest (QoI). We want to estimate the following probability
\[
\ell=\Pr(\BY\in B)=\mathbb{E}[I_B],
\]
where $I_B(\cdot)$ is an indicator function such that $I_B(\By)=1$ if $\By\in B$, and 0 otherwise.  To make our target problem more specific, we introduce the following two assumptions:
\begin{enumerate}
	\item The random variable $\BY$ can be effectively sampled.
	\item The probability $\mathbb{E}[I_B]$ is not too small. 
\end{enumerate}
These two assumptions simply mean that we are able to obtain a moderate number of effective samples satisfying $g(u)\geq0$ by directly sampling $\BY$. We can then consider the Monte Carlo estimator:
\begin{equation}
P_{\mathsf{MC}}=\frac{1}{N}\sum_{i=1}^NI_B(\By^{(i)}).
\end{equation}

To sample $u$, equation \eqref{eqn:general_model} is usually solved numerically. Let $\mathcal{L}_{h,f}$ and $\mathcal{L}_{h,c}$ indicate a fine and a coarse discretization of $\mathcal{L}$, where $h$ indicates a discretization parameter such as the element size in the finite element method. Let $u_{h,f}(t,\Bx,\BY)$ and $u_{h,c}(t,\Bx,\BY)$ be the two approximate solutions induced by $\mathcal{L}_{h,f}$ and $\mathcal{L}_{h,c}$ respectively. Since each sample of $\BY$ corresponds to solving a PDE, it can be very expensive if only $\mathcal{L}_{h,f}$ is employed for sampling. Then $\mathcal{L}_{h,c}$ is often used for variance reduction such that less samples from the fine model are needed to reach a certain accuracy, e.g., the multi-level Monte Carlo method \cite{Giles_AN15}. In this work, we consider a predictor-corrector strategy, which is widely used in scientific computing: 
\begin{enumerate}
	\item \textbf{Predictor}: We sample the reduced-order model to obtain the distribution of data satisfying $I_{B_{h,c}}=1$, where $B_{h,c}$ indicates the approximation of $B$ by $\mathcal{L}_{h,c}$.  
	\item \textbf{Corrector}: We correct the prediction given by the reduced-order model by sampling the fine model.
\end{enumerate}
The reasoning of the predictor-corrector strategy is as follows. The predictor given by the reduced-order model is relatively cheap to sample. Although the Monte Carlo estimator based on the reduced-order model is biased due to the error from dimension reduction, it provides useful information for variance reduction, meaning that the corrector based on the fine model does not require a large number of samples. In the next section, we will give a detailed presentation of the predictor-corrector strategy in the framework of importance sampling. 
\begin{remark}
In this work, we will not take into account the error of $\mathcal{L}_{h,f}$. When we say the reduced-order model induces a biased estimator, the bias is up to the accuracy of the fine model. 
\end{remark}

\subsection{Importance sampling}
Let $\rho(\By)$ be the probability density function of $\BY$. The basic idea of importance sampling is to compute the expectation with respect to another density function $\eta(\By)$ such as
\begin{equation}
\ell=\int I_B(\By)\frac{\rho(\By)}{\eta(\By)}\eta(\By)d\By=\mathbb{E}_\eta\left[
I_B(\By)\frac{\rho(\By)}{\eta(\By)}.
\right]
\end{equation}
The corresponding estimator is 
\begin{equation}
\hat{\ell}=\frac{1}{N}\sum_{i=1}^NI_B(\By)W(\By^{(i)}),
\end{equation}
where $W(\By)=\rho(\By)/\eta(\By)$ is the likelihood ratio. It is well known that the best candidate for the change of measure is 
\begin{equation}
\eta^*(\By)=\frac{I_B(\By)\rho(\By)}{\ell},
\end{equation}
i.e., the conditional PDF of $\By$ satisfying $I_B(\By)=1$. For this case, we have 
\[
\frac{I_B(\By^{(i)})\rho(\By^{(i)})}{\eta^*(\By^{(i)})}=\ell,
\]
meaning that the variance of this estimator is zero, where the superscript $*^{(i)}$ is the index for samples. Since $\ell$ is unknown, $\eta^*(\By)$ is only of theoretical importance. 

In reality, we usually replace $\eta^*(\By)$ with an approximate one, which is, among a family of parameterized PDFs, the closest one to the data set $\{\By^{(i)}|I_B(\By^{(i)})=1\}$.  In our problem setting, extra difficulties come from the fact that each sample $\By^{(i)}$ corresponds to solving a PDE, which can be time consuming. One commonly used strategy to alleviate this difficulty is to take advantage of the reduced-order model to achieve a trade-off between efficiency and accuracy. To estimate $\eta^*$, two cases can be considered depending on the source of the data: (1) The data  are just from the reduced-order model, or (2) The data are from both the reduced-order model and the fine model. For simplicity, we will only consider the first case in this work, where we need to resolve the following two general issues:
\begin{enumerate}
	\item How to use the data from the reduced-order model $\mathcal{L}_{h,c}$ to estimate $\eta^*$? A straightforward way is to approximate data distribution given by $B_{h,c}=\{\By^{(i)}|g(u_{h,c})\geq 0\}$. The problem of doing this is that the data satisfying $g(u_{h,c})\geq0$ may not satisfy $g(u_{h,f})\geq0$ due to the approximation errors of model reduction. In other words, $\eta^*$ is not absolutely continuous to its approximation. 
	\item How to choose a model $\eta(\By;\Btheta)$ for the density estimation, where $\Btheta$ indicates the model parameter. A widely used model is the Gaussian mixture, which can be viewed as a kind of kernel method. It is well known that learning high-dimensional Gaussian mixtures is difficult due to the curse of dimensionality. 
\end{enumerate} 

\subsection{Our general methodology}
Corresponding to the aforementioned two general issues, our methodology consists of two parts: 1) data preparation, where we include some extra data that satisfy $g(u_{h,c})<0$ and define a weighted empirical distribution, and 2) density estimation, where we resort to {machine} learning to construct an explicit model $\eta(\By;\Btheta)$. 

Before a detailed presentation of our methodology, we generalize the understanding of $\eta^*(\By)$ for the change of measure in the importance sampling. The effectiveness of the importance sampling estimator is determined by the variance of the function
\begin{equation}
w(\BY)=\frac{I_B(\BY)\rho(\BY)}{\eta(\BY)}.
\end{equation}
When $w(\BY)$ provides an unbiased estimator, i.e., $\mathbb{E}_{\eta}[w]=\mathbb{E}_{\rho}[I_B]$, the effectiveness of the estimator is determined by the second-order moment of $w(\BY)$:
\begin{equation}
\mathbb{E}_{\eta}[w^2]=\int_B\frac{\rho^2}{\eta}d\By.
\end{equation}
Due to the introduction of reduced-order model, we cannot guarantee that all data from the reduced-order model satisfy $g(u_{h,f})\geq 0$. Instead we can assume that the density estimation will be implemented on a set $\hat{B}$ that is larger than $B$, i.e., $B\subset \hat{B}$. This means that 
\begin{equation}\label{eqn:B_a}
\int_{B}\eta(\By)d\By=\alpha<1.
\end{equation}
We now look for the best $\eta$ which satisfies equation \eqref{eqn:B_a}, and minimizes the second-order moment of $w$. In other words, we consider the optimization problem 
\[
\min_{\eta}\left[J(\eta)=\int_B\frac{\rho^2}{\eta}d\By+\lambda\left(\int_B\eta d\By-\alpha\right)\right],
\]
where $\lambda$ is a Lagrange multiplier. Considering the first-order variation, we have  
\[
\delta J=-\int_B\frac{\rho^2}{\eta^2}\delta \eta d\By+\lambda\int_B\delta \eta d\By, 
\]
where $\delta \eta$ is a perturbation function. This means that the optimal $\eta$ satisfies
\begin{equation}\label{eqn:constant_ratio_B}
\frac{\rho^2}{\eta^2}=\lambda,\quad\forall\By\in B,
\end{equation}
from which we obtain the minimizer
\begin{equation}
\eta_\alpha^*(\By)=\frac{\alpha}{\mathbb{E}[I_B]}\rho(\By),\quad\forall \By\in B.
\end{equation}
The value of $\eta^*_\alpha$ on $\hat{B}\backslash B$ does not affect the performance of $\eta_\alpha^*$. 
The variance of $I_B$ is 
\begin{equation}
\mathrm{Var}(I_B)=\mathbb{E}[I_B]-\mathbb{E}[I_B]^2.
\end{equation}
The variance of $w(\BY)$ is 
\begin{equation}\label{eqn:vr_alpha}
\mathrm{Var}(w)=\frac{1}{\alpha}\mathbb{E}[I_B]^2-\mathbb{E}[I_B]^2.
\end{equation}
Thus, the closer $\alpha$ is to 1, the smaller the variance of $w$ is. When $\alpha=1$, i.e., $\hat{B}=B$, we have the best scenario with zero variance. Note that $\mathrm{Var}(w)>\mathrm{Var}(I_B)$ if $\alpha<\mathbb{E}[I_B]$. 

To this end, we obtain the following two general principles to {guide the development of} our methodology: 1) $\eta$ must provide a substantial probability on $B$, i.e., $\alpha$ should be close to 1; and 2) On $B$, the ratio between $\eta^*_\alpha$ and $\rho$ is always a constant even if $\eta_\alpha^*$ has a larger support than $B$.

\section{Change of measure via generative models}\label{sec:generative_model}

\subsection{Flow-based generative models}
Density estimation is a difficult problem especially for high-dimensional data. Many techniques have recently been developed in the framework of {machine} learning under the term generative modeling. Generative models are usually with likelihood-based methods, such as the autoregressive models \cite{Graves_2013,Oord_2016a,Oord_2016b}, variational autoencoders \cite{Kingma_2016}, and flow-based generative models \cite{Dinh_2014,Dinh_2016,Dhariwal_2018}. A particular case is the generative adversarial networks (GANs) \cite{Goodfellow_2014}, which requires finding a Nash equilibrium of a game. All generative models rely on the ability of deep nets for the nonlinear approximation of high-dimensional mapping. To incorporate the generative modeling into our problem setting, we here pay particular attention to the flow-based generative model. Simply speaking, the flow-based generative model implements a change of variable though an invertible mapping. It has two distinct features: 1) it provides an explicit form of the probability density function (PDF), and 2) it is easy to sample the estimated distribution. Other generative models usually do not have these two features at the same time. For example, GANs do not require an explicit form of the PDF, which makes it very flexible, but not straightforward for our purpose. 

Let $\BY\in\mathbb{R}^n$ be a random variable associated with the given data. Our target is to estimate the PDF of $\BY$ using the available data. Consider another random variable $\BZ=f(\BY)\in\mathbb{R}^n$, where $f(\cdot)$ is a bijection: $f:\BY\mapsto \BZ$. Let $p_{\BY}$ and $p_{\BZ}$ be the PDFs of $\BY$ and $\BZ$, respectively. We have
\begin{equation}\label{eqn:pdf_model}
p_{\BY}(\By)=p_{\BZ}(f(\By))\left|\det\nabla_{\By} f\right|.
\end{equation}
Once a prior distribution $p_{\BZ}(\Bz)$ is specified for $\BZ$, equation \eqref{eqn:pdf_model} provides a model for the density estimation of $\BY$. The key component of this model is the nonlinear mapping $f(\cdot)$. In flow-based generative models, an invertible mapping $f(\cdot)$ is constructed by deep nets. After the density estimation, the samples of $\BY$ can be easily generated as $\BY=f^{-1}(\BZ)$, thanks to the invertible mapping. 

To construct $f(\cdot)$, the main difficulties are twofold: (1) $f(\cdot)$ is highly nonlinear since the prior distribution for $\BZ$ must be simple enough, and (2) the mapping $f(\cdot)$ is a bijection. Flow-based generative models deal with these difficulties by stacking together a sequence of simple bijections, each of which is a shallow neural network, and the overall mapping is a deep net.  Mathematically, the mapping $f(\cdot)$ can be written in a composite form:
\begin{equation}\label{eqn:multi-layer-mapping}
\Bz=f(\By)=f_{[L]}\circ\ldots\circ f_{[1]}(\By),
\end{equation}
where $f_{[i]}$ indicates a coupling layer at stage $i$. The mapping $f_{[i]}(\cdot)$ is expected to \xw{be} simple enough such that its inverse and Jacobi matrix can be easily computed. Then given any $\Bz$, we can efficiently compute the inverse
\begin{equation}
\By=f^{-1}(\Bz)=f_{[1]}^{-1}\circ\ldots\circ f_{[L]}^{-1}(\Bz).
\end{equation}
Using the chain rule of differentiation, the determinant of the Jacobian matrix is obtained as 
\begin{equation}
|\det\nabla_{\By}f|=\prod_{i=1}^L|\det\nabla_{\By_{[i-1]}}f_{[i]}|,
\end{equation}
where $\By_{[i-1]}$ indicate the intermediate variables with $\By_{[0]}=\By$ and $\By_{[L]}=\Bz$.

One way to define $f_{[i]}$ is given by the real NVP \cite{Dinh_2016}. Consider a partition 
$\BY=(\BY_1,\BY_2)$ with $\BY_1\in \mathbb{R}^m$ and $\BY_2\in\mathbb{R}^{n-m}$. A simple bijection $f_{[i]}$ is  defined as
\begin{align}
\Bz_1&=\By_1,\label{eqn:affine_1}\\
\Bz_2&=\By_2\odot \Bs(\By_1)+\Bt(\By_1),\label{eqn:affine_2}
\end{align}
where $\Bs$ and $\Bt$ stand for scaling and translation depending only on $\By_1$, and $\odot$ indicates the Hadamard product or component-wise product. Note that only part of the input vector is updated using the information that depends on the rest of the input vector. The inverse of this mapping is also simple:
\begin{align}
\By_1&=\Bz_1,\\
\By_2&=(\Bz_2-\Bt(\By_1))/\Bs(\By_1),
\end{align}
where the division is component-wise. Note that the mappings $\Bs(\By_1)$ and $\Bt(\By_1)$ can be arbitrarily complicated, which will be modeled as a neural network (NN), i.e., 
\begin{equation}\label{eqn:NN}
(\log \Bs, \Bt) = \textsf{NN}(\By_1).
\end{equation}
The simple bijection given by equations \eqref{eqn:affine_1} and \eqref{eqn:affine_2} is also referred to as an affine coupling layer \cite{Dinh_2016}. Since only part of the input vector is updated, several affine couple layers need to be stacked together to update the whole input vector. 
The Jacobian matrix induced by one affine coupling layer is lower triangular:
\begin{equation}
\nabla_{\By}\Bz=\left[
\begin{array}{cc}
\BI&0\\
\nabla_{\By_1}\Bz_2&\textrm{diag}(\Bs(\By_1))
\end{array}
\right],
\end{equation}
whose determinant can be easily computed as 
\begin{equation}
\log|\det \nabla_{\By}\Bz|=\sum_{i=1}^{n-m}\log|s_i(\By_1)|.
\end{equation}

\subsection{Improve the multi-layer invertible mapping $f(\cdot)$}
It is seen that the multi-layer invertible mapping $f(\By)$ relies on the stacking of some simple coupling layers $f_{[i]}$. For the effectiveness of this strategy, we need to pay attention to several issues.
\subsubsection{The depth $L$} If $\By$ is partitioned to two parts, at least two affine coupling layers are needed for a complete modification of $\By$. Note that the modification of  $\By_2$ in equation \eqref{eqn:affine_2} is linear, meaning that a large depth $L$ may be needed such that enough correlations between $\By_1$ and $\By_2$ are introduced.  It usually is enough to define a shallow neural network $\mathsf{NN}$ for each affine coupling layer since the update given by $f_{[i]}$  is limited by its definition. In this work, we use two fully coupled hidden layers for $\mathsf{NN}$ {(see equation \eqref{eqn:NN})}. The capability of $f(\By)$ mainly relies on the depth $L$.
 
\subsubsection{The partition of $\BY$} We have several different choices for the partition of $\BY$:
	\begin{enumerate}
		\item Fixed partition. This is the choice we are using so far for the presentation. In every affine coupling layer, the first $m$ components are modified or remain the same, where we usually let $m=\lfloor n/2\rfloor$. The drawback of this choice is twofold: (1) we treat the two halves of $\BY$ equally although they may not be of the same importance; and (2) The degree of mixing of all components of $\BY$ is limited. For example, if $\By_1$ is nearly independent of $\By_2$, we expect to mix the components of $\By_1$ instead of modifying $\By_1$ linearly using a function of $\By_2$. 
		\item Random partition. If we do not have {a} prior knowledge of the importance of each dimension of $\BY$, a random partition provides a simple way to increase the correlation between the components of $\BY$.  {The random partition shuffles all the components of $\By$ before implementing a fixed partition such that each coupling layer $f_{[i]}$ has a different partition pattern.} 
		\item Linear transformation of $\BY$. We define a new random variable $\hat{\BY}=\mathbf{W}{\BY}$, where $\mathbf{W}$ is a non-singular matrix and can be regarded as a rotation between two coordinate systems. We then consider a fixed partition of $\hat{\BY}$ instead of $\BY$. Furthermore, we include $\mathbf{W}$ into the trainable parameters. In other words, although we do not know the importance of each dimension of $\BY$ for the desired nonlinear mapping, we can let the algorithm learn from the data a better coordinate system for the fixed partition. The most important dimension may be given by a linear combination of $\BY$. So the optimization of $\mathbf{W}$ acts like principle component analysis (PCA). In \cite{Dhariwal_2018} a similar  strategy was used to improve the performance of real NVP for image processing. 
	\end{enumerate}
In this work, we mainly stick to the fixed partition of $\BY$ to test the effectiveness of our methodology. Once the effectiveness is verified, the second and third options can be employed for further improvement. 	
	
\subsubsection{Scale and bias layer} It is well known that batch normalization can improve the propagation of training signal in a  deep net. Let ${\tilde{\boldsymbol{\mu}}}$ and $\tilde{\boldsymbol{\sigma}}^2$ be the mean and variance estimated from the mini batch \cite{Szegedy_2015}. The batch normalization algorithm includes two steps: the first step defines for each layer of the neural network the following normalization
	\begin{equation}\label{eqn:batch_normalization_mean_std}
	y_i\leftarrow\frac{y_i-\tilde{\mu}_i}{\sqrt{\tilde{\sigma}_i^2+\epsilon}},\quad i=1,\ldots,n,
	\end{equation}
	and the second step refines the previous step by a trainable scale-shift operation:
	\begin{equation}
	\hat{\By}=\gamma\By+\boldsymbol{\beta}.
	\end{equation}	
When the size of minibatch is small, batch normalization  \eqref{eqn:batch_normalization_mean_std} becomes less effective due to the noise in the compuration of $\tilde{\boldsymbol{\mu}}$ and $\tilde{\boldsymbol{\sigma}}$. A compromise of the two steps in the batch normalization algorithm is proposed in \cite{Dhariwal_2018}, i.e.,
	\begin{equation}\label{eqn:actnorm}
	\hat{\By} = \Ba\odot\By+\Bb, 
	\end{equation}
where $\Ba$ and $\Bb$ are trainable, and initialized by $\tilde{\boldsymbol{\mu}}$ and $\tilde{\boldsymbol{\sigma}}$ associated with the initial data. After the initialization, $\Ba$ and $\Bb$ will be treated as regular trainable parameters that are independent of the data. In this work, we simplify the procedure \eqref{eqn:actnorm} further by only applying the scale and bias layer given by equation \eqref{eqn:actnorm} to the input of $f_{[i]}$. In other words, we do not apply any normalization techniques to the shallow neural network for $\Bs(\cdot)$ and $\Bt(\cdot)$. The only motivation of this simplification is to study the robustness of the generative model in our problem setting. 

\begin{figure}
	\center{
		\includegraphics[width=0.4\textwidth]{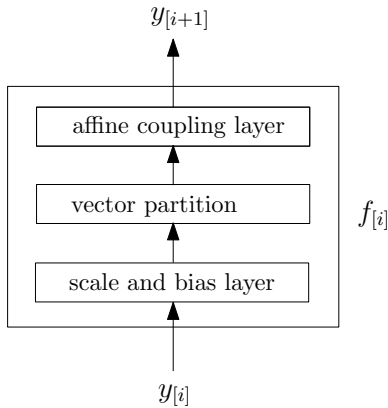}
	}
	\caption{The diagram of a general coupling layer $f_{[i]}$.}\label{fig:2}
\end{figure}

Combining the above discussions, we can refine the coupling layer $f_{[i]}$ as shown in figure \ref{fig:2}, where the input of an affine coupling layer is partitioned in a certain way after a scale and shift layer is implemented.

\section{Cross entropy by the weighted empirical distribution}\label{sec:wed}
\subsection{Likelihood and cross entropy}
The generative model will be trained by maximizing the likelihood. In terms of data, the minimum cross entropy is the same as the maximum likelihood. Consider the set of data $\{\By^{(i)}\}_{i=1}^N$ from the distribution of $\BY$. Specifying a distribution for $\BZ$ in equation \eqref{eqn:pdf_model}, we obtain a model for the PDF of $\BY$. Let $\theta$ be the parameter from the definition of the mapping $f(\By)$. The maximum likelihood estimator of $\theta$ is 
\begin{equation}
\theta_{\textsf{ML}}=\mathrm{argmax}_\theta\prod_{i=1}^Np_{\BY}(\By^{(i)};\theta)=\mathrm{argmax}_\theta \sum_{i=1}^N\log p_{\BY}(\By^{(i)};\theta).
\end{equation}
Let ${\mu}_{\textsf{data}}(\By)$ be the empirical distribution of the data. Multiplying $1/N$ to the right-hand side of the above equation, the maximum likelihood estimator can also be regarded as
\begin{equation}\label{eqn:ML}
\theta_{\textsf{ML}}=\mathrm{argmax}_\theta \mathbb{E}_{{\mu}_{\textsf{data}}}[\log p_{\BY}(\By;\theta)],
\end{equation}
where the expectation is with respect to ${\mu}_{\textsf{data}}$. To estimate $\theta$, we can also minimize the distance between ${\mu}_{\textsf{data}}$ and ${\mu}_{\BY}$ using the Kullback-Leibler (KL) divergence:
\begin{equation}
D_{\textsf{KL}}({\mu}_{\textsf{data}}||\mu_{\BY})=\mathbb{E}_{{\mu}_{\textsf{data}}}\left[
\log\frac{d{\mu}_{\textsf{data}}}{d\mu_{\BY}}
\right]=H(\mu_{\textsf{data}},\mu_{\BY})-H(\mu_\textsf{data})
\end{equation}
where $\mu_{\BY}(d\By)=p_{\BY}d\By$, $H(\mu_{\textsf{data}},\mu_{\BY})$ is the cross entropy of $\mu_{\textsf{data}}$ and $\mu_{\BY}$, and $H(\mu_{\textsf{data}})$ is the entropy of the empirical distribution solely determined by the data. It is seen that to minimize the KL divergence, we only need to minimize the cross entropy
\begin{equation}\label{eqn:crossentropy}
H(\mu_{\textsf{data}},\mu_{\BY})=-\mathbb{E}_{{\mu}_{\textsf{data}}}[\log p_{\BY}]=-\frac{1}{N}\sum_{i=1}^N\log p_{\BY}(\By^{(i)};\theta),
\end{equation}
because the entropy $H(\mu_{\textsf{data}})$ only depends on data. Comparing equations \eqref{eqn:ML} and \eqref{eqn:crossentropy}, we know that maximizing the maximum likelihood is equivalent to minimizing the cross entropy. Let us assume that the components of {$\BZ$} are i.i.d. normal random variables. We then have
\begin{align*}
\log p_{\BY}(\By)& = \log|\det\nabla_{\By}f|-\frac{1}{2}\sum_{i=1}^nz_i^2(\By)-d\log\sqrt{2\pi}\\
&=\sum_{i=1}^L\log|\det\nabla_{\By_{[i-1]}}f_{[i]}|-\frac{1}{2}\sum_{i=1}^nz_i^2(\By)-d\log\sqrt{2\pi}.
\end{align*}

\subsection{Weighted empirical distribution}
We still consider the set of data $\{\By^{(i)}\}_{i=1}^N$ from the distribution $\mu_{\BY}$ of $\BY$. The empirical measure $\mu_N$ associated with the data set is defined as 
\begin{equation}
\mu_N(A)=\frac{1}{N}\sum_{i=1}^NI_A(\By^{(i)})=\frac{1}{N}\sum_{i=1}^N
\delta_{\By^{(i)}}(A),
\end{equation} 
where $\delta_{\By}$ is the Dirac measure. We also define a weighted version of $\mu_N$ as follows:
\begin{equation}
\hat{\mu}_N(A)=\sum_{i=1}^Nw_i\delta_{\By^{(i)}}(A),
\end{equation}
with $\sum_{i=1}^Nw_i=1$. It recovers the empirical measure when $w_i=\frac{1}{N}$. For the empirical measure, each sample in the data set is equally important in the sense of the law of large numbers, since all samples have the same weight $1/N$ and are obtained independently. However, in reality we are often more interested in the information of $\BY$ that satisfies a certain constraint. A simple and flexible way to incorporate constraints into the data is to associate the data with varying weights. 

Let us consider a simple scenario to illustrate the weighted empirical measure. We partition data set $\{\By^{(i)}\}_{i=1}^N=\{\By^{(i)}\}_{i\in\mathcal{I}_1}\cup\{\By^{(i)}\}_{i\in\mathcal{I}_2}$ with $\mathcal{I}_1\cap\mathcal{I}_2=\emptyset$, where $\mathcal{I}_i$ indicates an index set with $i=1,2$. We let
\[
w_i=\pi_1,\,\forall i\in\mathcal{I}_1\quad\textrm{ and }\quad w_i=\pi_2,\,\forall i\in\mathcal{I}_2,
\]
where $\pi_1,\pi_2\geq0$ are two constants, satisfying $N_1\pi_1+N_2\pi_2=1$ with $N_i$ being the cardinality of $\mathcal{I}_i$, $i=1,2$. We expect to emphasize the information given by the data set $\{\By^{(i)}\}_{i\in\mathcal{I}_1}$ by increasing the value of $\pi_1$. For simplicity, we assume that  $\{\By^{(i)}\}_{i\in\mathcal{I}_1}\subset A\subset\mathbb{R}^n$, and 
$\{\By^{(i)}\}_{i\in\mathcal{I}_2}\subset A^c$ with $A^c$ being the complement of $A$. Let  
\[
\rho_1(\By)=\frac{\rho_{\BY}(\By)}{\mathbb{E}_{\mu_{\BY}}[I_A]},\quad
\rho_2(\By)=\frac{\rho_{\BY}(\By)}{\mathbb{E}_{\mu_{\BY}}[I_{A^c}]}
\]
be the two conditional PDFs. We then seek a PDF of the form
\begin{equation}
\rho_w(\By;\gamma)=\gamma\rho_1(\By)+(1-\gamma)\rho_2(\By),
\end{equation}
that is closest to the weighted measure $\hat{\mu}_N$ in terms of the KL divergence, where $0<\gamma<1$. For the given data set, minimizing the KL divergence is equivalent to minimizing the cross entropy 
\begin{align}
H(\hat{\mu}_N,\eta)&=\mathbb{E}_{\hat{\mu}_N}\left[\log\rho_w(\By)\right]\\
&=\sum_{i\in\mathcal{I}_1}\pi_1\log(\rho_w(\By^{(i)};\gamma))+\sum_{i\in\mathcal{I}_2}\pi_2\log(\rho_w(\By^{(i)};\gamma)),\nonumber
\end{align}
where $\eta(d\By)=\rho_w d\By$. Then $\partial_\gamma H=0$ yields that
\begin{equation}
\frac{N_1\pi_1}{\gamma}-\frac{N_2\pi_2}{1-\gamma}=0\quad\Rightarrow\quad \gamma=N_1\pi_1.
\end{equation}
When $\pi_1=1/N$ as in the empirical distribution, $\gamma=\frac{N_1}{N}\approx\mathbb{E}_{\mu_{\BY}}[I_{A}]$.  It is seen that if we increase the weights for the data in $\{\By^{(i)}\}_{i\in\mathcal{I}_1}$, the corresponding PDF $\rho_w(\By)$ will increases the probability of taking values in $A$, compared to the PDF $\rho_{\BY}(\By)$. 

When considering the weighted empirical distribution, we only need a slight modification  of the objective function for the minimization of the cross entropy, where equation \eqref{eqn:crossentropy} becomes
\begin{equation}\label{eqn:crossentropy_weighted}
H(\hat{\mu}_{N},\mu_{\BY})=-\mathbb{E}_{\hat{\mu}_{N}}[\log p_{\BY}]=-\sum_{i=1}^Nw_i\log p_{\BY}(\By^{(i)};\theta),
\end{equation}
which corresponds to the maximization of a weighted likelihood:
\begin{equation}
\prod_{i=1}^N p_{\BY}^{\gamma_i}(\By^{(i)};\theta)
\end{equation}
with $\gamma_i=Nw_i$.

\subsection{Weight the data given by the reduced-order model}
Recall that the optimal choice for the change of measure in importance sampling is 
\[
\eta^*(\By)=\frac{I_{B}(\By)\rho(\By)}{\ell}.
\]
Sampling the reduced-order model, a straightforward approximation of $\eta^*(\By)$ is 
\begin{equation}
\eta^*_{h,c}(\By)=\frac{I_{B_{h,c}}(\By)\rho(\By)}{\ell_{h,c}},
\end{equation}
where $\ell_{h,c}=\mathbb{E}[I_{B_{h,c}}]$. Due to the errors induced by model reduction,  $\eta^*(\By)$ is not absolutely continuous with respect to $\eta_{h,c}^*(\By)$. More specifically, when $I_{B_{h,c}}=0$ or $\eta_{h,c}^*(\By)=0$, it is possible that $I_B=1$, i.e., $\eta^*(\By)>0$. If $\eta_{h,c}^*(\By)$ is used for importance sampling, the estimation will be obviously biased, although the convergence can still be reached as the numerical discretization of $u$ is refined. An easy way to fix this problem is to enlarge the support of $\eta^*_{h,c}$ by incorporating the error estimate of $g(u_{h,c})$. Note that for any $\By$, we have
\begin{align*}
g(u_{h,c})=g(u)+\left\langle\frac{\delta g}{\delta u},u_{h,c}-u\right\rangle+\mathit{O}(\|u_{h,c}-u\|^2),
\end{align*}
where $\frac{\delta g}{\delta u}$ indicates the functional derivative and $\langle\cdot,\cdot\rangle$ the inner product in the physical space. The first-order variation of $g(u)$ in terms of $u-u_{h,c}$ yields the leading term in the error of $g(u_{h,c})$. Instead of $g(u_{h,c})=0$, we can obtain a better guess of $g(u)=0$ using
\[
g(u_{h,c})\approx 0+\left\langle\frac{\delta g}{\delta u},u_{h,c}-u\right\rangle,
\]
which is possibly smaller than 0. When sampling the reduced-order model, we need to keep the data satisfying
\begin{equation}\label{eqn:g_1st_variation}
g(u_{h,c})\geq -\left|\left\langle\frac{\delta g}{\delta u},u_{h,c}-u\right\rangle\right|,
\end{equation}
such that we will not reject the data satisfying $g(u_{h,c})<0$ while $g(u)\geq 0$. In reality, the error of $g(u_{h,c})$ can be estimated by a posterior error estimate techniques, which has a general form
\begin{equation}
|g(u_{h,c})-g(u)|\leq C_{\By}h^m,
\end{equation}
where $C_{\By}$ is a positive constant depending on $\By$, and $m$ is an index indicating the accuracy of the reduced-order model. Instead of using equation \eqref{eqn:g_1st_variation}, we can use 
\begin{equation}\label{eqn:g_acceptance}
g(u_{h,c})\geq -Ch^m
\end{equation}
as the acceptance criterion of data, where $C$ is a positive constant chosen according to $C_{\By^{(i)}}$.  Unfortunately, by doing this, we often accept a lot of redundant data. We rewrite 
\[
g(u_{h,c}) = g(u)+\epsilon(\By)
\]
and look at the discrepancy between $I_{\{g(u_{h,c})\geq0\}}$ and $I_{\{g(u)\geq0\}}$. Note that $\{g(u_{h,c})\geq0\}=\{g(u)\geq-\epsilon(\By)\}$.  If $\epsilon\geq0$, $\{g(u)\geq0\}\subseteq\{g(u_{h,c})\geq0\}$; if $\epsilon\leq0$, $\{g(u_{h,c})\geq0\}\subseteq\{g(u)\geq0\}$. 
Thus the information missed  by $I_{\{g(u_{h,c})\geq0\}}$ is that $\{0\leq g(u)<-\epsilon(\By)\}$ subject to the condition that $\epsilon(\By)\leq 0$. In terms of $u_h$, what is missing is that $\{\epsilon(\By)\leq g(u_{h,c})<0\}$ when $\epsilon(\By)\leq 0$. Then the following data included by equation \eqref{eqn:g_acceptance}, are unnecessary:
\begin{align*}
\left\{
\begin{array}{ll}
\{g(u_{h,c})<0\}, & \textrm{ if }\epsilon(\By)\geq0,\\
\{-Ch^\alpha \leq g(u_{h,c})<-\epsilon(\By)\}, & \textrm{ if }\epsilon(\By)\leq0.
\end{array}\right.
\end{align*}
The portion of the redundant data in $\{-C h^\alpha\leq g(u_{h,c})<0\}$ is 
\begin{align*}
&\frac{\Pr(\{-C h^m\leq g(u_{h,c})<0\})\Pr(\epsilon\geq0)
+\Pr(\{-C h^m\leq g(u_h)<-\epsilon\})\Pr(\epsilon\leq0)}{\Pr(\{-C h^m\leq g(u_{h,c})<0\})}\\
=&\Pr(\epsilon\geq 0)+\frac{\Pr(\{-C h^m\leq g(u_{h,c})<-\epsilon\})\Pr(\epsilon\leq0)}{\Pr(\{-C h^m\leq g(u_{h,c})<0\})}\\
\approx&\frac{1}{2}+\frac{1}{2}\frac{\Pr(\{-C h^m\leq g(u_{h,c})<-\epsilon\})}{\Pr(\{-C h^m\leq g(u_{h,c})<0\})},
\end{align*}
where we assume that $\Pr(\epsilon\leq 0)\approx\Pr(\epsilon\geq0)\approx\frac{1}{2}$. In other words, at least $50\%$ of the data in $\{-C h^m\leq g(u_{h,c})<0\}$ are not necessary, and  if the a posterior error estimate is not tight, most of the data are redundant. If $\mathbb{E}[I_{\{g(u_{h,c})>0\}}]$ is relatively small, the scenario is worse since the probability induced by the unnecessary data might be larger than $\mathbb{E}[I_{\{g(u_{h,c})>0\}}]$. For this case, most of the data may be nothing but pollution (see the example in section \ref{sec:elliptic_1d}) in terms of the approximation of $\eta^*(\By)$. According to equation \eqref{eqn:vr_alpha}, a large amount redundant data implies a small $\alpha$ which makes it difficult to achieve variance reduction. 

\begin{figure}
	\center{
		\includegraphics[width=0.8\textwidth]{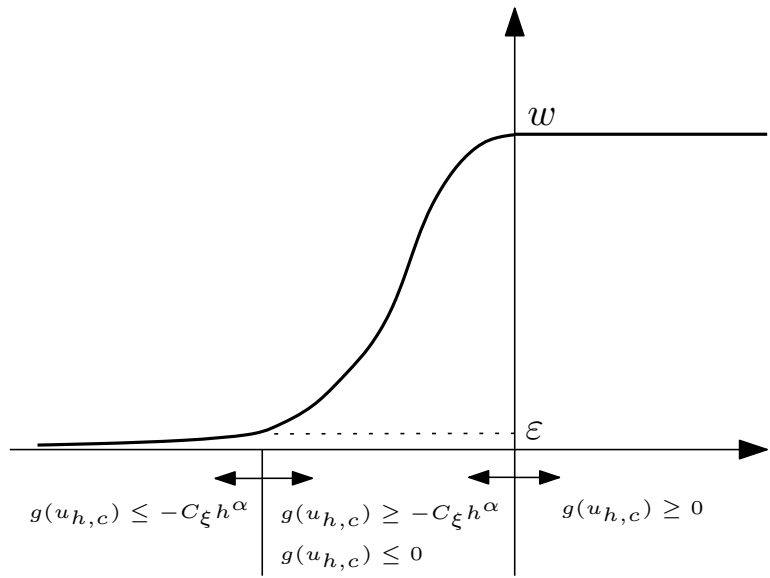}
	}
	\caption{The diagram of the weight distribution for the data points from sampling the coarse model.}\label{fig:wed}
\end{figure}

To deal with this issue, we will adjust the weights of the data such that the undesired data do not contribute too much in the empirical distribution. Following is our plan to weight the data from the reduced-order model:
\begin{itemize}
	\item All data points satisfying $g({u_{h,c}})\geq0$ share the same weight, which mimics equation \eqref{eqn:constant_ratio_B}.
	\item For the data points satisfying $g(u_{h,c})\in[-Ch^m, 0]$, the weight decreases exponentially as $g(u_{h,c})$ decreases away from 0, as shown in figure \ref{fig:wed}. We will use a half-normal distribution in terms of $g(u_{h,c})$ to weight the data.     
\end{itemize}
More details about the implementation will be given in section \ref{sec:implementation}.

\subsection{A penalty term}
Let ${\eta}^*_{h,c}(\By)$ be the estimated PDF using the weighted data from the reduced-order model. If the generative model is overly complex or the size of data set for training is not large enough, we need to pay particular attention to  the overfitting. Many general techniques such as early {stopping} have been developed in {machine} learning \cite{Zhang_AS05}.  We here focus on regularization related to our problem setting. If the conditional pdf $\eta^*(\By)$ can be well approximated, we should expect that 
\begin{equation}\label{eqn:weight_constant}
w_{h,c}(\By)=\frac{I_{B}(\By)\rho(\By)}{{\eta}^*_{h,c}(\By)}\approx C, 
\end{equation}
or 
\begin{equation}
\nabla_{\By}\log w_{h,c}(\By)\approx0,
\end{equation}
where $C$ is a positive constant. If the overfitting is not a concern, equation \eqref{eqn:weight_constant} is a natural result given by the density estimation. However, when the overfitting occurs, the minimization of the cross entropy $H(\mu_{\mathsf{data}},\mu_{\BY})$ may yield an approximate distribution such that $w_{h,c}$ has a large standard deviation, which means that the important sampling estimator based on ${\eta}_{h,c}^*$ may fail to induce variance reduction. 

To increase the robustness of the algorithm, we want to balance the minimization of the cross entropy and the condition \eqref{eqn:weight_constant}. A convenient way to do this is to add a penalty term 
\begin{align}
&\beta\mathbb{E}_{p_{\BY}}\left[\left|\nabla_{\By}\log w_{h,c}(\By)\right|^2\right]^{1/2}\nonumber\\
=&\beta\left(\int \left|\rho^{-1}\nabla\rho-p_{\BY}^{-1}\nabla p_{\BY}\right|^2p_{\BY}d\By\right)^{1/2},
\end{align}
{into the objective function}, where $\beta$ is a penalty parameter. The term $\nabla p_{\BY}$ in the integrand provides a $H_1$ regularization of the objective function. Note that the condition $\By\in B$ is determined by the fine model, which is unknown. In reality, we compute the penalty term  with respect to the weighted empirical distribution, i.e., 
\begin{equation}\label{eqn:penalty_empirical}
{\beta}\mathbb{E}_{\hat{\mu}_N}\left[|\nabla_{\By}\log w_{h,c}(\By)|^2\right]^{1/2}.
\end{equation}
To this end, we have the final objective function for training the generative model as
\begin{equation}
H(\hat{\mu}_N,\mu_{\BY})+{\beta}\mathbb{E}_{\hat{\mu}_N}\left[|\nabla_{\By}\log w_{h,c}(\By)|^2\right]^{1/2},
\end{equation}
where $\mu_{\BY}(d\By)=p_{\BY}d\By$.

\section{Implementation}\label{sec:implementation}

We sample $\BY$ to obtain $\{\By^{(i)}\}_{i=1}^M$. For each $\By^{(i)}$, we solve a PDE to obtain $u_{h,c}(\By^{(i)})$, and compute an error estimate $\epsilon_{h,c}(\By^{(i)})$ of $g(u_{h,c}(\By^{(i)})$. Let $g_{h,c}(\By)=g(u_{h,c}(\By))$. We organize the data as $\{(\By^{(i)},\epsilon_{h,c}(\By^{(i)}),g_{h,c}(\By^{(i)}))\}_{i=1}^M$. Let
\[
\epsilon^{-}_{\mathrm{max}}=\max_{i}\left|\epsilon_{h,c}(\By^{(i)})I_{\{g_{h,c}(\By^{(i)})<0\}}\right|.
\]
We will keep the data $\{(\By^{(i)},\epsilon_{h,c}(\By^{(i)}),g_{h,c}(\By^{(i)}))\}_{i=1}^N$, where $g_{h,c}(\By^{(i)})\geq -\epsilon^{-}_{\mathrm{max}}$. 
We then use the half-normal distribution 
\[
f_{\tau}(\tau;\sigma)=\frac{\sqrt{2}}{\sigma\sqrt{\pi}}\exp\left(-\frac{\tau^2}{2\sigma^2}\right),\quad z\geq0
\]
to fit the data $\tau^{(i)}=g_{h,c}(\By^{(i)})$ satisfying $-\epsilon_{\mathrm{max}}^-\leq g_{h,c}(\By^{(i)})<0$. For the data $\By^{(i)}$, we associate a weight 
\begin{equation}
w_i=\left\{
\begin{array}{ll}
c_1f_\tau(\tau^{(i)})),&\text{ if }\tau^{(i)}<0,\\
c_2,&\text{ if }\tau^{(i)}\geq 0,
\end{array}
\right.
\end{equation}
where $c_1$ and $c_2$ are two positive constants. Let $N_+$ be the number of $\By^{(i)}$ satisfying $g_{h,c}(\By^{(i)})\geq 0$. We determine $c_1$, $c_2$ and $\sigma$ using the following relations:
\begin{equation}\label{eqn:wgt}
\left\{
\begin{array}{rcl}
N_+c_2&=&\theta,\\
c_1\sum_{i=1}^{N-N_+}f_{\tau}(\tau^{(i)})&=&1-\theta,\\
c_1\frac{\sqrt{2}}{\sigma\sqrt{\pi}}&=&c_2,
\end{array}
\right.
\end{equation}
where $0<\theta<1$. We assign uniform weights to the data $\{g_{h,c}(\By^{(i)})\geq 0\}$, whose probability from the weighted empirical distribution is $\theta$. The data $\{g_{h,c}(\By^{(i)})< 0\}$ has a probability $1-\theta$, where the weight decays exponentially as the value $|g_{h,c}(\By)|$ increases. The third equation can be regarded as a continuity condition, meaning that weight should be continuous when crossing the interface $g_{h,c}(\By)=0$. It is seen that $c_2$ can be easily obtained from the first equation. From the third equation, we have $c_1=\frac{c_2\sigma\sqrt{\pi}}{\sqrt{2}}$, which simplifies the second equation as
\[
\sum_{i=1}^{N-N_+}c_2\exp\left(-\frac{(\tau^{(i)})^2}{2\sigma^2}\right)=1-\theta.
\]
Note that the left-hand side is an increasing function with respect to $\sigma\in(0,+\infty)$, meaning there exists a unique $\sigma\in(0,+\infty)$ satisfying the above equation. Considering $\sigma=\alpha\max_i|g_{h,c}(\By^{(i)})|$, we have 
\[
\sum_{i=1}^{N-N_+}c_2\exp\left(-\frac{(\tau^{(i)})^2}{2\sigma^2}\right)>\theta\left(\frac{N}{N_+}-1\right)\exp\left(-\frac{1}{2\alpha^2}\right).
\]
Letting
\[
\theta(\frac{N}{N_+}-1)\exp\left(-\frac{1}{2\alpha^2}\right)=(1-\theta),
\]
i.e.,
\[
\alpha=\left(-0.5\left(\log\frac{1-\theta}{Nc_2-\theta}\right)^{-1}\right)^{1/2},
\]
we have the root located in $[0,\alpha\max_{i}|g_{h,c}(\By^{(i)})|]$, which can be computed numerically by a root-finding algorithm.

Another implement issue is related to the stochastic optimization. For unweighted data, a commonly used strategy in stochastic optimization is to split the uniformly shuffled training samples into small batches within one epoch. For the  weighted data, a uniform shuffle is obviously not optimal. We then generate batches in a way that is more consistent with the distribution of the weights. We partition the interval $[-\epsilon_{\mathrm{max}}^-,0]=\cup_{k=1}^Ke_k$ uniformly into $K$ disjoint sub-intervals $e_k$. Let $e_{K+1}=[0,\infty)$. We then group all the training samples as 
\[
S_k=\{\By^{(i)}|g_{h,c}(\By^{(i)})\in e_k\},\quad k=1,\ldots,K+1.
\]
We will shuffle the training samples in $S_k$ uniformly before we split each $S_k$ to a certain number of batches. We pick one batch in each $S_k$ to assemble the training batch for each iteration step of the stochastic optimization. 

Once the generative model $p_{\BY}(\By)$ is trained, we use it to construct an importance sampling estimator for the fine model
\begin{align}
\ell=\int I_{\{g(u_{h,f})\geq 0\}}\rho(\By)d\By&=\mathbb{E}_{p_{\BY}}\left[
I_{\{g(u_{h,f})\geq0\}}\frac{\rho(\By)}{p_{\BY}(\By)}
\right]\nonumber\\
&=\mathbb{E}_{p_{\BZ}}\left[
I_{\{g(u_{h,f})\geq0\}}\frac{\rho(f^{-1}(\Bz))}{p_{\BY}(f^{-1}(\Bz))}\right],
\end{align}
where $p_{\BZ}$ is the prior distribution, e.g., the Gaussian $\mathcal{N}(0,I)$ with $I$ being a $n$-dimensional identity matrix.

\section{Numerical experiments}\label{sec:num}
In this section, we do some experiments to study the numerical strategies we have proposed. The ADAM optimization solver with a fixed learning rate is used for all examples. 

\subsection{Rotate Gaussian random variables}
	We start with a simple case. Assume that we have data for the random variable $\BY=(Y_1,Y_2)$ with $Y_i$ being i.i.d. normal random variables. The entropy of $\mu_{\BY}$ is $H(\mu_{\BY})=\ln(2\pi e)$. We know that $\hat{\BY}=A\BY$ are still Gaussian random variables, where $A\in\mathbb{R}^{2\times2}$. Furthermore, 
	\[
	\mathrm{Cov}(\hat{\BY})=A\mathrm{Cov}(\BY)A^\mathsf{T}=AA^T.
	\]
	If $A$ is a unitary matrix, $\hat{Y}_1$ and $\hat{Y}_2$ are i.i.d. normal random variables. We use the flow-based generative model to describe the mapping, i.e., rotation, from $\hat{\BY}$ to $\BY$.  
	
\begin{figure}
	\center{
		\includegraphics[width=0.8\textwidth]{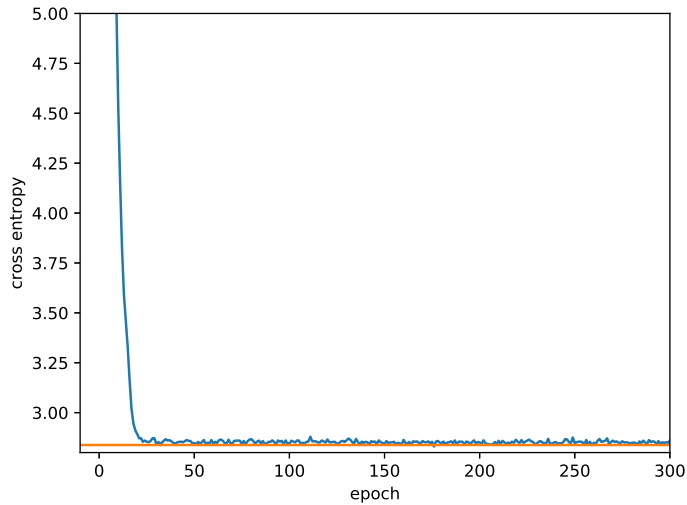}
	}
	\caption{The convergence behavior for the rotation of Gaussian variables, where the horizontal line indicates the entropy $H(\mu_{\BY})=\ln(2\pi e)$. Four general coupling layers are used, i.e., $L=2$. In equation \eqref{eqn:affine_2}, we let $\Bs(\cdot)=1$ and only model $\Bt(\cdot)$ as a nueral network $\mathsf{NN}(\cdot)$. The sample size is $N=10^4$.}\label{fig:rotate_gaussian}
\end{figure}

	Let us see if the multi-layer mapping $f(\Bx)$ defined in \eqref{eqn:multi-layer-mapping} is able to provide a rotation of $(Y_1,Y_2)$ using two affine coupling layers.  According to equations \eqref{eqn:affine_1} and \eqref{eqn:affine_2}, $f_{[1]}$ yields 
	\[
	y_1^{[1]}=y_1,\quad y_2^{[1]}=ay_2+by_1, 
	\]
	where we choose $s(y_1)=a$ and $t(y_1)=by_1$ with $a$, $b$ being constant. Similarly, we have the output of $f_{[2]}$ as 
	\[
	\hat{y}_1 = cy_1+d(ay_2+by_1)=(c+bd)y_1+ady_2, \quad \hat{y}_2=ay_2+by_1,
	\] 
	where two more constants $c$ and $d$ are introduced. We then obtain the following condition such that $A$ is unitary:
	\[
	\left(
	\begin{array}{cc}
	c+bd&ad\\
	a&b
	\end{array}
	\right)^\mathsf{T}
	\left(
	\begin{array}{cc}
	c+bd&ad\\
	a&b
	\end{array}
	\right)=
	\left(
	\begin{array}{cc}
	1&0\\
	0&1
	\end{array}
	\right)
	\]
	The above equation admits many possible solutions, e.g., $a=b=\sqrt{2}/2$, $c=-\sqrt{2}$ and $d=1$. Any possible solution is a good enough for our purpose. In equation \eqref{eqn:crossentropy}, $\mu_{\mathsf{data}}$ is given by $N$ samples of $\BY$. Then the cross entropy $H(\mu_{\mathsf{data}},\mu_{\BY})$ should converge to the entropy $H(\mu_{\BY})$, i.e., $\ln(2\pi e)\approx2.8379$,  as $N\rightarrow\infty$. If the flow-based generative model $p_{\BY}d\By=\tilde{\mu}_{\BY}$ provides a good approximation of $\mu_{\BY}$, the minimum of the cross entropy $H(\mu_{\mathsf{data}},\tilde{\mu}_{\BY})$ should yield a minimizer that converges to $\mu_{\BY}$ and a minimum value that converges to $H(\mu_{\BY})$. Such a convergence behavior is shown in figure \ref{fig:rotate_gaussian}, meaning that a rotation of Gaussian variables is well captured. The initial cross entropy is large because we choose a large standard deviation on purpose when we initialize the weights of each neuron. It is seen that the ADAM method stabilizes quickly. 

\subsection{Two-dimensional conditional PDFs}
\begin{figure}
	\center{
		\includegraphics[width=0.8\textwidth]{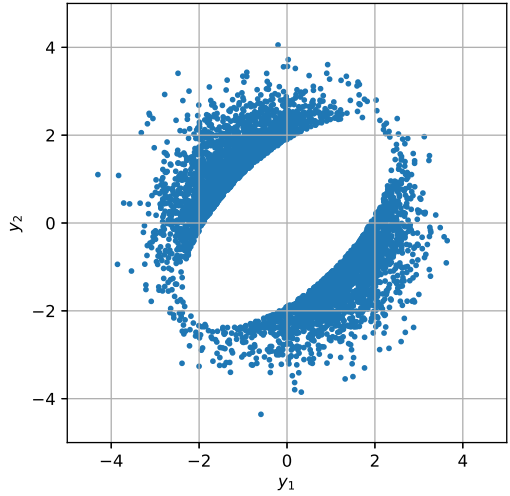}
	}
	\caption{The set $B=\{\By|\hat{\By}^\mathsf{T}\hat{\By}\geq C^2\}$ with $\alpha=2$, $\theta=\pi/4$ and $C=3.0$. We assume that $Y_1$ and $Y_2$ are two i.i.d. normal random variables.}\label{fig:2d_stars} 
\end{figure}

We now consider the approximation of the following conditional PDF 
\[
p_{\BY|B}(\By)=\frac{I_B(\By)\rho(\By)}{\mathbb{E}[I_B]},
\]
where we choose $\rho(\By)$ as the joint PDF given by two i.i.d. normal random variables $Y_1$ and $Y_2$. The condition $B=\{\By|g(\By)\geq 0\}$ will introduce correlations between $Y_1$ and $Y_2$.  Let  $\hat{\By}=\Lambda R\By$, where $\Lambda=\mathrm{diag}(\alpha, 1)$ is a scaling matrix with $\alpha$ being a constant, and $R$ is a unitary matrix for rotation, i.e.,
\[
R=\left[
\begin{array}{cc}
\cos\theta& -\sin\theta\\
\sin\theta&\cos\theta
\end{array}
\right].
\]
We define the set $B=\{\By|\hat{\By}^\mathsf{T}\hat{\By}\geq C^2\}$. The distribution of $p_{\BY|B}(\By)$ is demonstrated in figure \ref{fig:2d_stars} for $\alpha=2$, $\theta=\pi/4$ and $C=3.0$ by $N=5000$ samples. These are the data we will use to train the generative model. 

\begin{figure}
	\center{
		\includegraphics[width=0.99\textwidth]{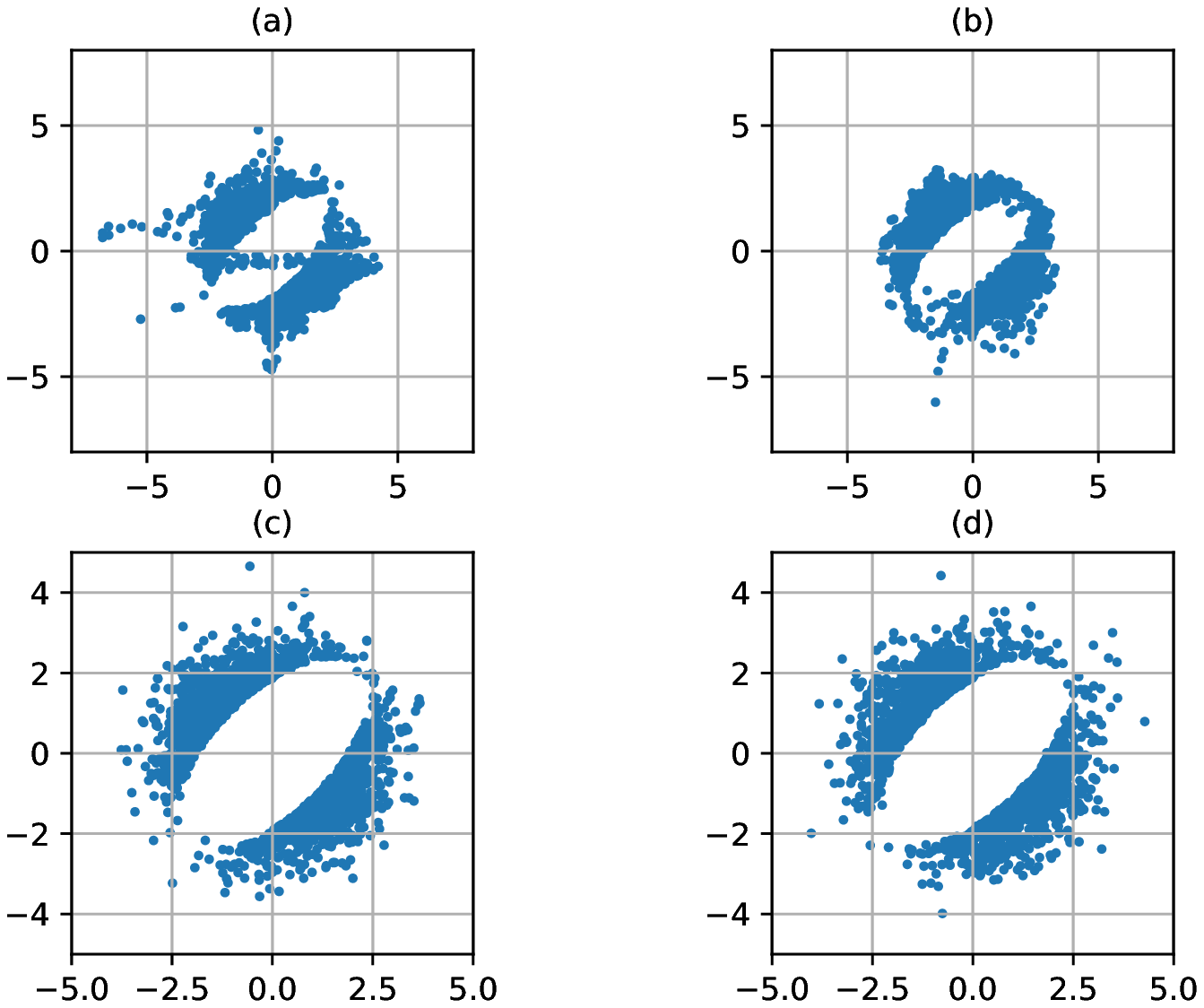}
	}
	\caption{Data sampled from the flow-based generative models with a Gaussian prior distribution. The sample size is $N=10^4$. (a): $L=2$; (b): $L=4$; (c): $L=8$; (d): $L=16$.  }\label{fig:prior_gaussian} 
\end{figure}

\begin{figure}
	\center{
		\includegraphics[width=0.99\textwidth]{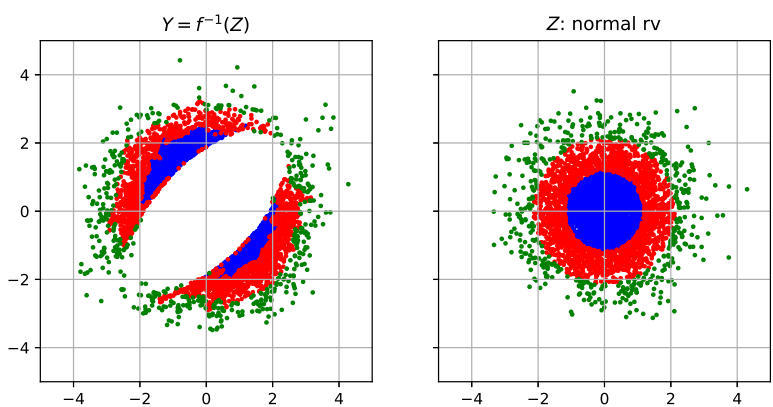}
	}
	\caption{The mapping from $\BZ$ to $\BY$ given by the generative model with $L=16$, where $\BZ$ is subject to the prior normal distribution $\mathcal{N}(0,I)$. The sample size is $N=10^4$. }\label{fig:transportation_prior_gaussian} 
\end{figure}

We let the prior distribution be the two-dimensional normal distribution $\mathcal{N}(0,I)$ with $I$ being a two-dimensional identity matrix. The transportation from the normal distribution to the desired conditional distribution is highly nonlinear due to the fact that the region of the highest density in the prior distribution has been removed. In figure \ref{fig:prior_gaussian}, we plot the data sampled from the generative models trained with different depths. The neural network $\mathsf{NN}(\cdot)$ in equation \eqref{eqn:NN} has two dense hidden layers, where the first hidden layer has 512 neurons and the second hidden layer has 256 neurons. It is seen that the approximated distribution improves as the depth $L$ increases. When $L=8$, the approximated distribution already agrees very well with the original distribution showed in figure \ref{fig:2d_stars}. In figure \ref{fig:transportation_prior_gaussian}, we demonstrate the mapping from $\BZ$ to $\BY$, where $\BZ$ is sampled from the Gaussian prior. For clarity, we split the data $\Bz$ to three groups, indicated by blue, red and green.   The one-to-one correspondence between $\Bz$ and $\By$ yields the corresponding splitting of the data $\By$. It appears that the nonlinear mapping $f(\cdot)$ overall maps the high-density region in the prior distribution to the high-density region in the data distribution. Note that the blue region has been separated into two parts, meaning that the deep net is able to handle such a ``discontinuity'' using a continuous mapping. 

\subsection{One-dimensional elliptic problems with log-normal coefficients}\label{sec:elliptic_1d}
We now consider a one-dimensional elliptic problem \cite{Wan_model3}
\begin{equation}
-\frac{d}{dx}\left(e^{a(x;\omega)}\frac{du}{dx}\right)=1,\quad x\in[0,1],
\end{equation}
where $a(x;\omega)$ is a zero-mean Gaussian random field subject to a normalized covariance kernel $K(x_1,x_2)$. For this one-dimensional problem, we can write down the exact solution
\begin{equation}
u(x;\omega)=-\int_0^xse^{-a(s;\omega)}ds+\gamma\int_0^xe^{-a(s;\omega)}ds,
\end{equation}
where $\gamma$ is a random variable
\[
\gamma=\left(\int_0^1e^{-a(s;\omega)}ds\right)^{-1}\int_0^1se^{-a(s;\omega)}ds.
\]
The random coefficient $a(x;\omega)$ can be approximated by the Karhunen-Lo\'{e}ve expansion:
\begin{equation}
a(x;\omega)\approx a_M(x;\xi)=\sigma\sum_{i=1}^M\sqrt{\lambda_i}\theta_i(x)\xi_i,
\end{equation}
where $\sigma$ indicates the standard deviation, $\xi_i\sim\mathcal{N}(0,1)$ are i.i.d. normal random variables, and $(\lambda_i,\theta_i(x)) $ are the eigen-pairs of the covariance kernel $K(x_1,x_2)$. $\sigma$ will be fixed to 1 from now on. Replacing $a(x)$ with $a_M(x)$ in $u$, we obtain $u_M(x)\approx u(x)$, which will be our exact solution.  Define the set $B=\{u_M|\|u_M\|_{H^1}\geq C\}$ with $C$ being a positive number. We will estimate $\mathbb{E}[I_B]$ by sampling.

Consider a one-dimensional exponential covariance kernel on $x\in[0,1]$
$$K(x_1,x_2)=e^{-\frac{|x_1-x_2|}{l_c}}.$$ 
Its eigenvalues satisfy
\begin{equation}
v^2=\frac{2\epsilon-\epsilon^2\lambda_i}{\lambda_i},\quad (v^2-\epsilon^2)\tan(v)-2\epsilon v=0,
\end{equation}
where $\epsilon=1/l_c$. Its eigenfuncitons have the following form \cite{Karnia_JSC02}
\begin{equation}
\theta_i(x)=\frac{v\cos(vx)+\epsilon\sin(vx)}{\sqrt{\frac{1}{2}(\epsilon^2+v^2)+(w^2-\epsilon^2)\frac{\sin(2v)}{4v}+\frac{\epsilon}{2}(1-\cos(2v))}}.
\end{equation} 
Let $\Pi_{h,c}$ be an interpolation operator defined on the coarse mesh. We let 
\[
a_{M,h,c}(x;\By)=\sum_{i=1}^M\sqrt{\lambda_i}\Pi_{h,c}\theta_i(x)\xi_i,
\]
which yields the approximate solution $u_{M,h,c}$. For the reduced-order model, all the integrals will be approximated by the rectangle rule which has a first-order accuracy. The fine model will be based on spectral/$hp$ element method. More specifically, we consider the interpolation and integration using 64 equidistant elements with 8 Gauss-Lobatto-Legendre points in each element. For simplicity, the error of the reduced-order  model will be computed directly using the fine model as the reference solution.

\subsubsection{Distribution of data missed by the reduced-order model}
\begin{figure}
	\center{
		\includegraphics[width=0.49\textwidth]{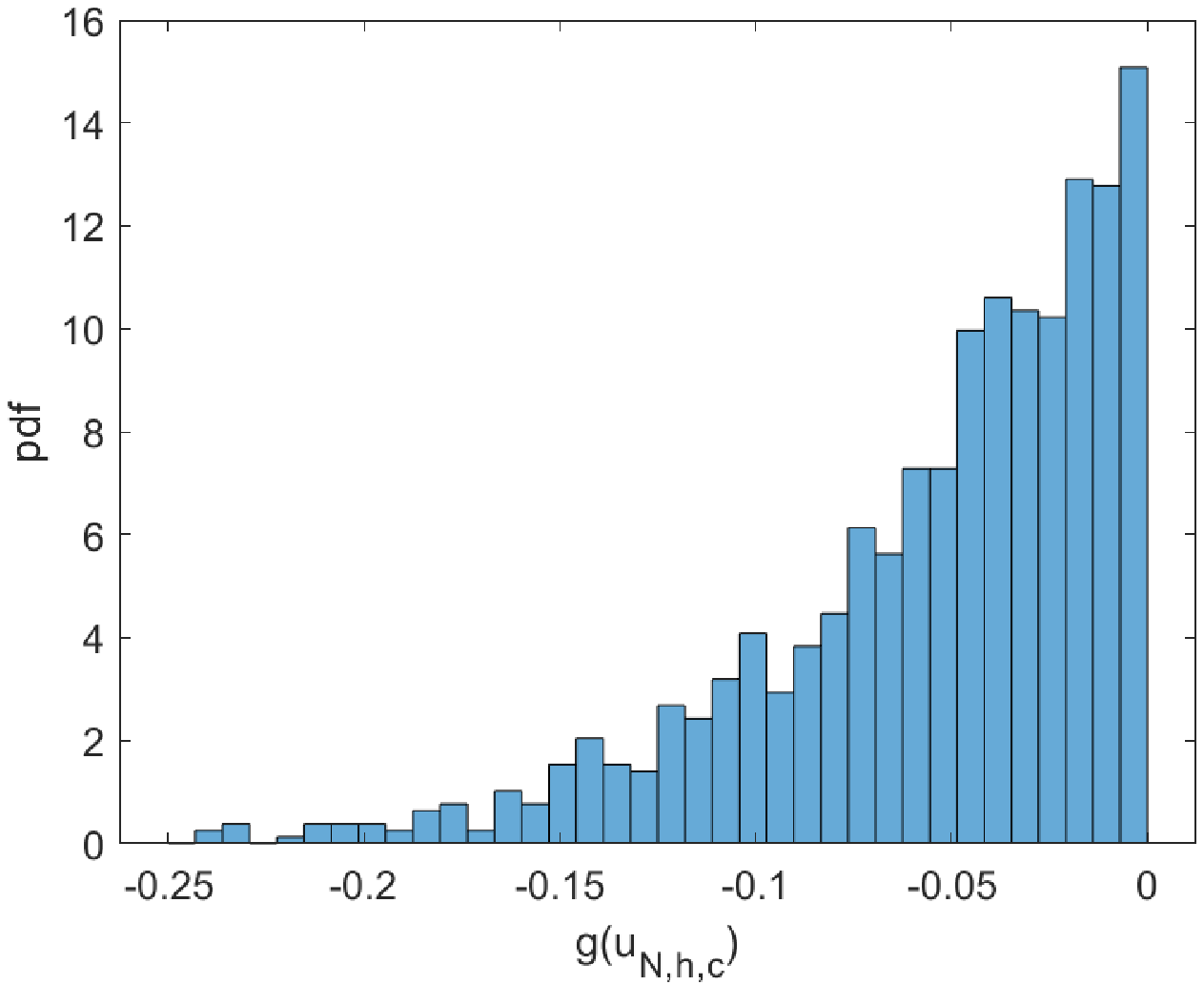}
		\includegraphics[width=0.49\textwidth]{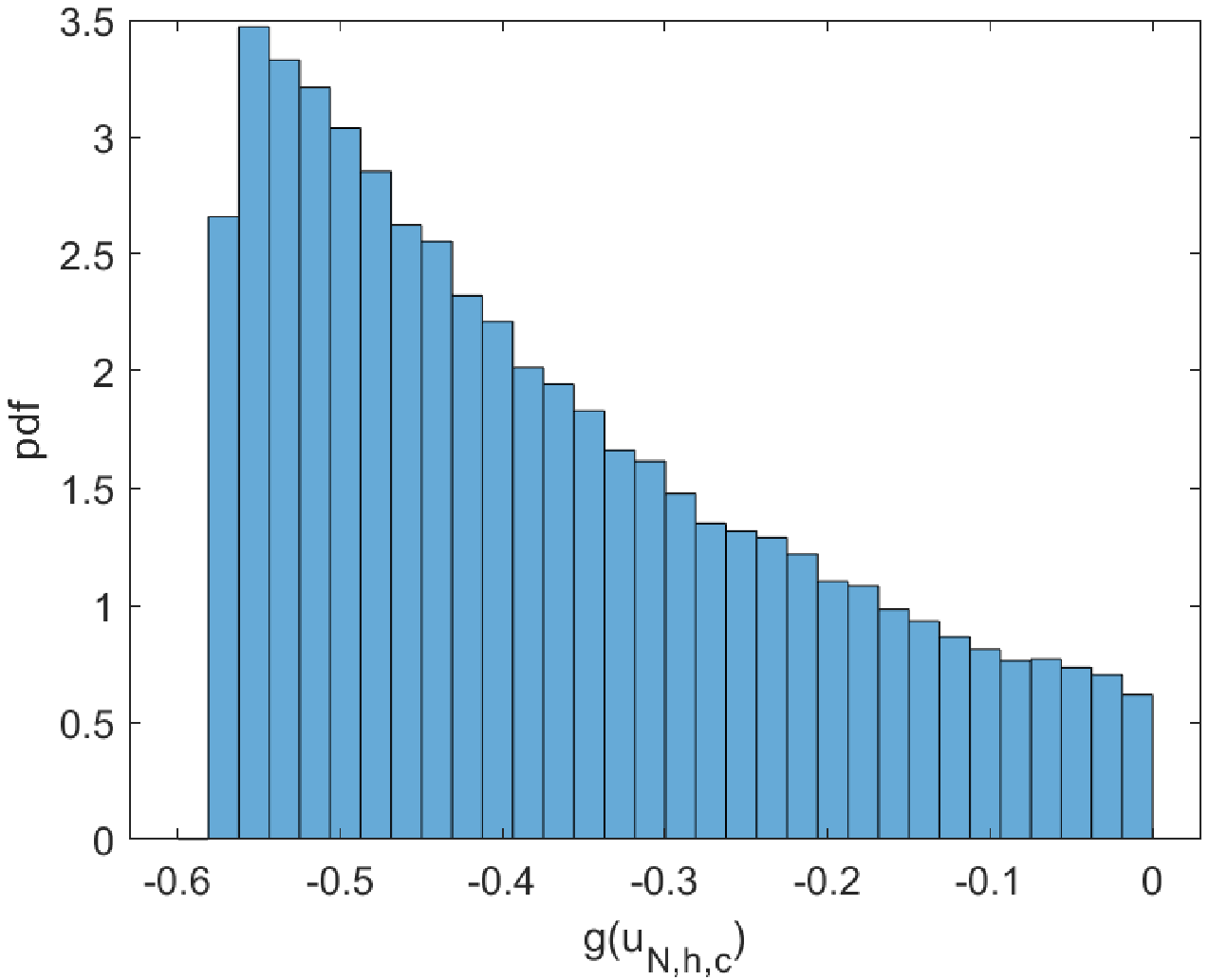}
	}
	\caption{The conditioned distribution of $g(u_{N,h,c})$. Left: The data missed by the coarse model, i.e., $g(u_{N,h,c})<0$ while $g(u_{N,h,f})\geq0$. Right: The data that satisfy 
	$C-\epsilon_{\mathrm{max}}^{-}\leq \|u_{N,h,c}\|_{H^1}< C$. }\label{fig:hist_missed_data} 
\end{figure}

We first look at the necessity of considering the weighted empirical distribution. In table \ref{tbl:compare_transition}, we summarize the information about $10^4$ samples from both the reduced-order and the fine models, where the mesh for the reduced-order model $u_{h,c}$ consists of 10 equidistant linear finite elements. The probability $\Pr(B)$ is chosen around 0.1. It is seen that to reduce the bias from the reduced-order model, we need to keep another 2,546 samples that do not satisfy $\|u_{M,h,c}\|_{H^1}\geq C$. However, among these samples, only 107 are effective, which is around $\frac{107}{2546}\approx4\%$. If we do density estimation using 1,263 + 2,546 = 3,809 samples, 2,546 - 107 = 2,539 samples do not contribute at all to our desired random event, which are $\frac{2,539}{3,809}\approx67\%$ of the total samples. Such a situation can be worse if we use a posterior error estimate because the effective index of the estimator may be several times larger than 1, i.e., the estimated error may be several times larger than the real error. To alleviate this issue, we need to put more weights into the 1,263 samples that satisfy $\|u_{M,h,c}\|_{H^1}\geq C$ and less weights to the redundant 2,546 samples that are induced by the discretization error of the reduced-order  model. We note that the 107 useful samples will also be weighed by doing so. A compromise is to assign the weights to the data $\{C-\epsilon^-_{\mathrm{max}}\leq\|u_{M,h,c}\|<C\}$ in a consistent way with the distribution of the data $\{\|u_{M,h,c}\|<C\textrm{ and }\|u_{M,h,f}\|\geq C\}$. 
In figure \ref{fig:hist_missed_data}, we plot the normalized histograms of some conditioned distribution of $g(u_{M,h,c})=\|u_{M,h,c}\|-C$. In the left plot of figure \ref{fig:hist_missed_data}, we show the distribution of $g(u_{M,h,c})$ given by the data where the reduced-order model fails to capture $B$, i.e., $g(u_{M,h,c})<0$ while $g(u_{M,h,f})\geq0$. It is seen that as the value of $g(u_{M,h,c})$ decreases, the probability that the reduced-order model fails also decreases. In the right plot of figure \ref{fig:hist_missed_data}, we show the distribution of $g(u_{M,h,c})$ given by the data that satisfy  $-\epsilon_{\mathrm{max}}^{-}\leq g(u_{M,h,c})< 0$. It is seen that the density increases as the value of $g(u_{M,h,c})$ decreases, which is the opposite of the histogram in the left plot. This is because we have kept redundant data to compensate the discretization error of the reduced-order model. First, the probability that $C-\epsilon_{\mathrm{max}}^{-}\leq \|u_{M,h,f}\|_{H^1}< C$ is much larger than the probability that $\|u_{M,h,c}\|_{H^1}< C$ and $\| u_{M,h,f}\|_{H^1}\geq C$. Second, $\epsilon_{\mathrm{max}}^-$ is not the optimal choice, which may be much larger than necessary. At this moment, we do not have a better understanding about the choice of the lower bound for $-\epsilon_{\mathrm{max}}^{-}\leq g(u_{M,h,c})< 0$.
\begin{table}
\caption{\label{tbl:compare_transition}Samples from the coarse model, where $C=0.8$, $l_c=1$, and  $M=50$.}
\centering
\begin{tabular}{|c|c|c|c|}
\hline
\# of samples & $10^4$
 \\ \hline
$\|u_{M,h,c}\|_{H^1}\geq C$& 1,263 \\ \hline
$\|u_{M,h,f}\|_{H^1}\geq C$& 1,300 \\ \hline
$C-\epsilon_{\mathrm{max}}^{-}\leq \|u_{M,h,c}\|_{H^1}< C$ & 2,546\\ \hline
$\|u_{M,h,c}\|_{H^1}< C$ and $\|u_{M,h,f}\|_{H^1}\geq C$ &  107\\ \hline
\end{tabular}
\end{table}  

\subsubsection{Importance sampling via the trained generative model}
We now look at the performance of the generative model for the importance sampling estimator. Let $\sigma_{I_B}$ and $\sigma_w$ be the standard deviation of $I_B$ and 
\[
w(\BY)=\frac{I_B(\BY)\rho(\BY)}{p_{\BY}(\BY)},
\]
where $p_{\BY}$ indicates the trained generative model. Let $N_{\mathsf{MC}}$ and $N_{\mathsf{IS}}$ be the sample size for the Monte Carlo estimator and the importance sampling estimator to achieve the same degree of confidence interval for the mean subject to a certain error. We know that 
\[
\frac{N_{\mathsf{IS}}}{N_{\mathsf{MC}}}\approx\left(\frac{\sigma_w}{\sigma_{I_B}}\right)^2.
\]
So we only need to focus on the variance reduction of $\sigma_w$ in terms of $\sigma_{I_B}$.

Following is the setup of our numerical experiments. We compute $\sigma_{I_B}$ using the fine physical model by the Monte Carlo method with $10^5$ samples. The depth $L$ of the generative model is set to $16$. Each affine coupling layer has two fully coupled hidden layers, where the first one has 512 neurons and the second one has 256 neurons. In each coupling layer $f_{[i]}$, we consider a fixed partition of the vector. Considering that the eigenvalue decays, we split $\boldsymbol{\xi}=(\xi_1,\xi_2,\ldots,\xi_{2m})$ into $\boldsymbol{\xi}_1=(\xi_1,\xi_3,\ldots,\xi_{2m-1})$ and $\boldsymbol{\xi}_2=(\xi_2,\xi_4,\ldots,\xi_{2m})$, where all components with odd indices are separated from those with even indices.  We then train the generative model using the data given by the reduced-order model and compute $\sigma_{w}$ by sampling the generative model $10^5$ times. For all cases, the generative model will be trained by the ADAM method with a learning rate 2e-4, where {the data have been split to 23 minibatches}. We sample the reduced-order model $10^5$ times, and keep a portion of the data as the training set. Since we choose that $\Pr(B)\approx 0.1$, about $10^4$ samples satisfy $g(u_{h,c})\geq 0$, although the real number may vary a little. We set $\theta=0.85$ when computing the weights of the data. 

\begin{figure}
	\center{
		\includegraphics[width=0.49\textwidth]{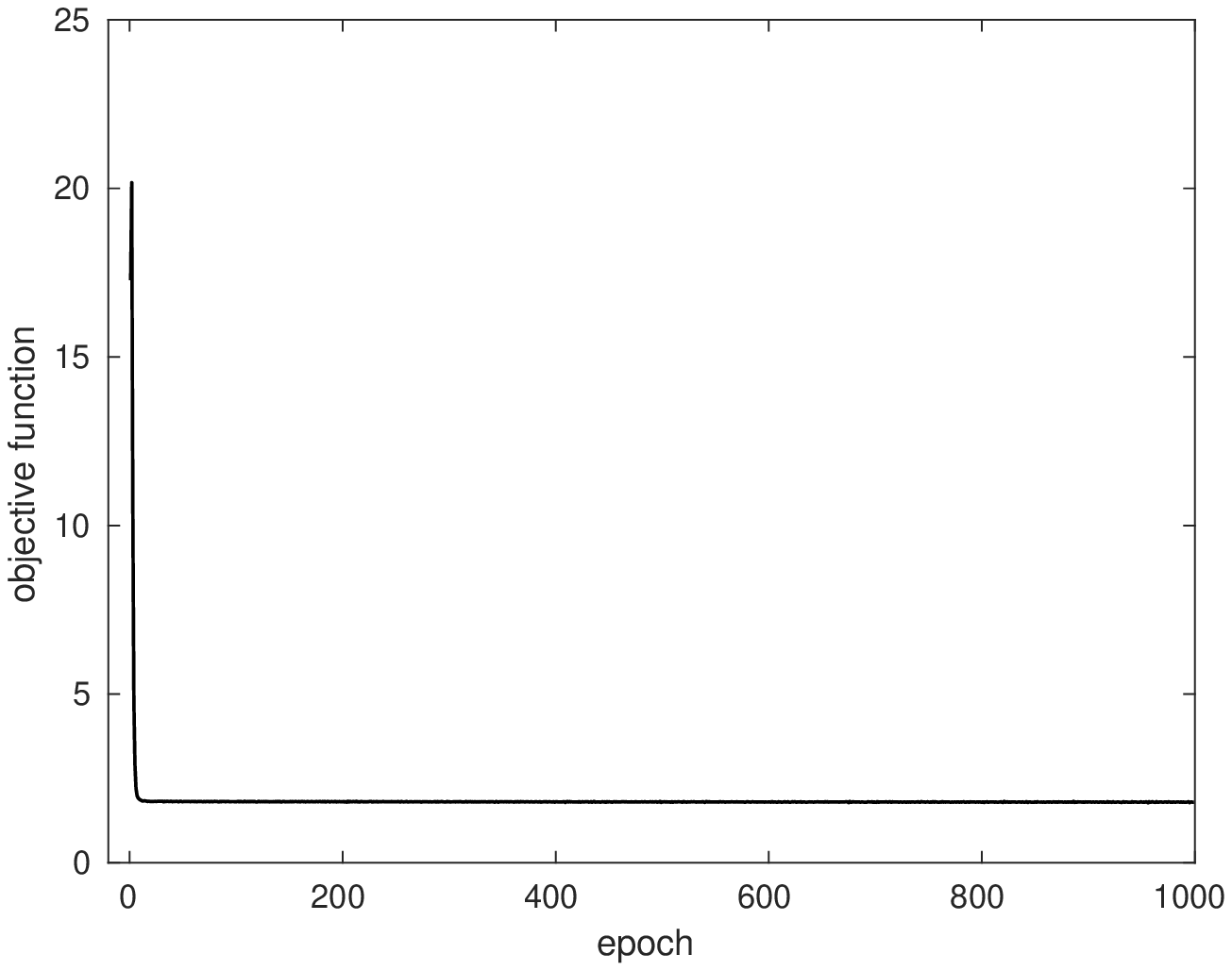}
		\includegraphics[width=0.49\textwidth]{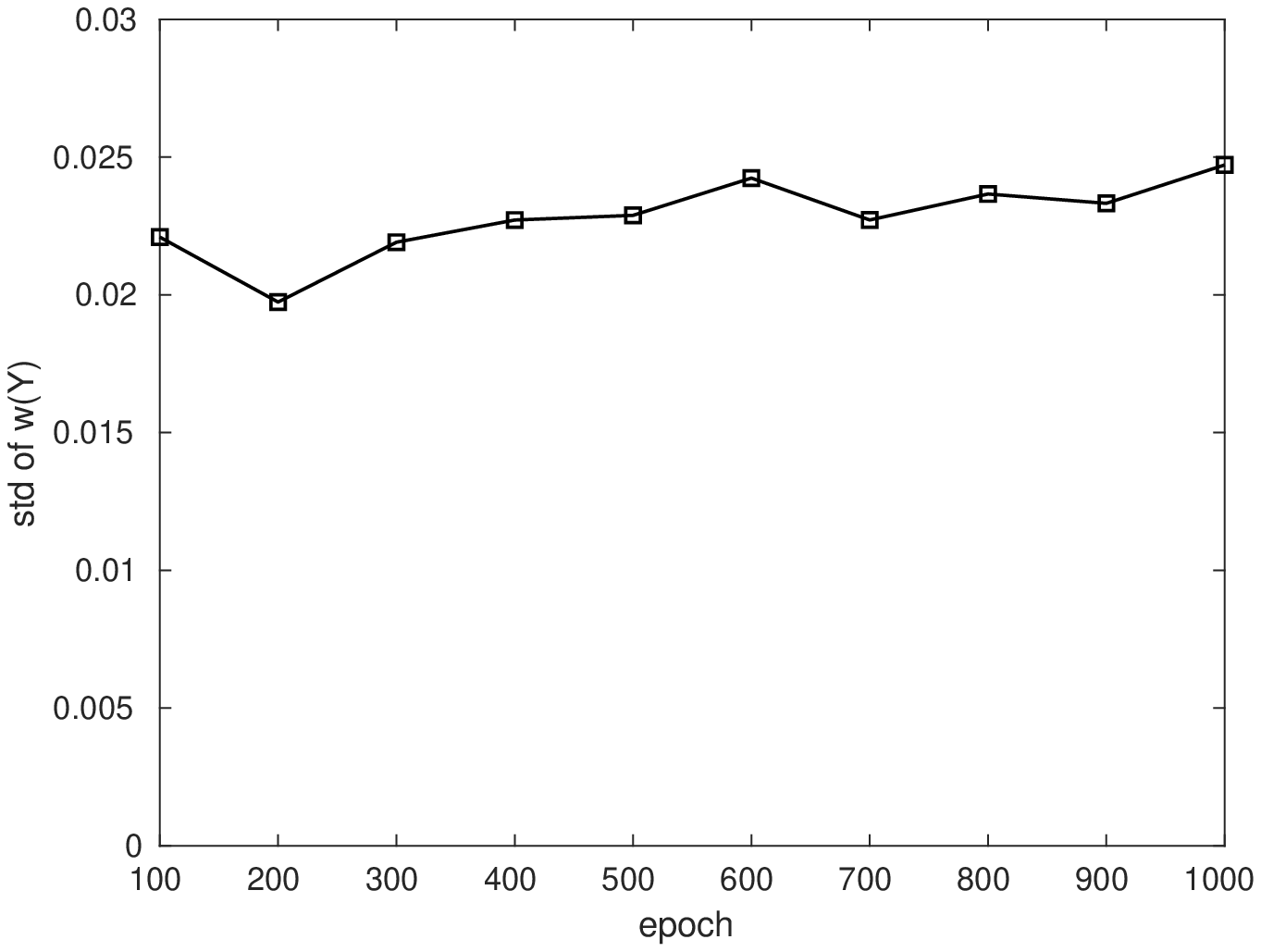}
	}
	\caption{$M=2$. Left: The evolution behavior of stochastic optimization, where the penalty term is not included in the objective function. Right: The standard deviation of $w(\BY)$.}\label{fig:d2_cong_std} 
\end{figure}

\begin{figure}
	\center{
		\includegraphics[width=\textwidth]{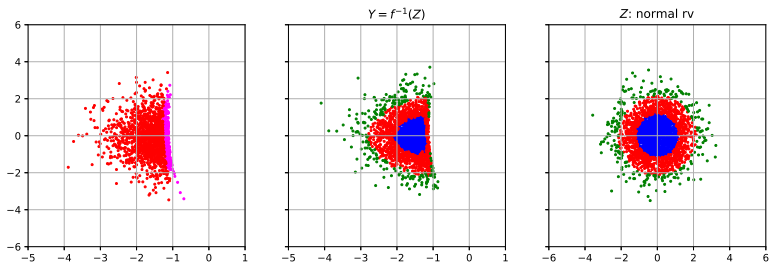}
	}
	\caption{Left: The data from the reduced-order models, where the red color indicates $g(u_{h,c})\geq 0$ while the megenta color indicates $g(u_{h,c})< 0$. Middle: The estimated distribution given by the generative model with $L=16$; Right: The priori distribution given by two iid normal random variables. }\label{fig:density_estimation_elliptic_N2} 
\end{figure}

We start with a relatively large correlation length $l_c=1$, such that the eigenvalue decays fast. The coarse mesh consists of 10 equidistant linear finite elements. We first look at a two-dimensional case, i.e., $M=2$, where $\mathbb{E}[I_B]\approx\textrm{0.109}$ and $\sigma_{I_B}\approx\textrm{0.312}$. The training set from the reduced-order model includes 10,683 samples satisfying $g(u_{M,h,c})\geq 0$, and 3,051 samples satisfying $g(u_{M,h,c})< 0$, among which only 242 samples are really missed by the reduced-order model, i.e., $g(u_{M,h,c})< 0$ while $g(u_{M,h,f})\geq 0$. In figure \ref{fig:d2_cong_std}, we plot the results for $M=2$. On the left, we plot the {evolution} behavior of the stochastic optimization, where no penalty term is included in the objective function, i.e., $\beta=0$; On the right, we plot the standard deviation of $\sigma_w$ versus the epoch, where $\sigma_w$ is computed in terms of the generative model {trained up to} a certain epoch. It is seen that the stochastic optimization stabilizes quickly while $\sigma_w$ varies a little {around} 0.025. For this case, 
\[
\frac{N_{\mathsf{IS}}}{N_{\mathsf{MC}}}\approx\left(\frac{0.025}{0.312}\right)^2\approx0.64\%.
\]
In other words, for the same level of accuracy, the number of samples needed by the  importance sampling estimator is about \%0.64 of that for a direct Monte Carlo estimator. The speed up can be significant even after taking into account the cost from sampling the reduced-order model and training the generative model, since the complexity of the generative model does not increase with the complexity of the physical model. The comparison between the data distribution and the estimated distribution is given in figure \ref{fig:density_estimation_elliptic_N2}.

\begin{figure}
	\center{
		\includegraphics[width=0.49\textwidth]{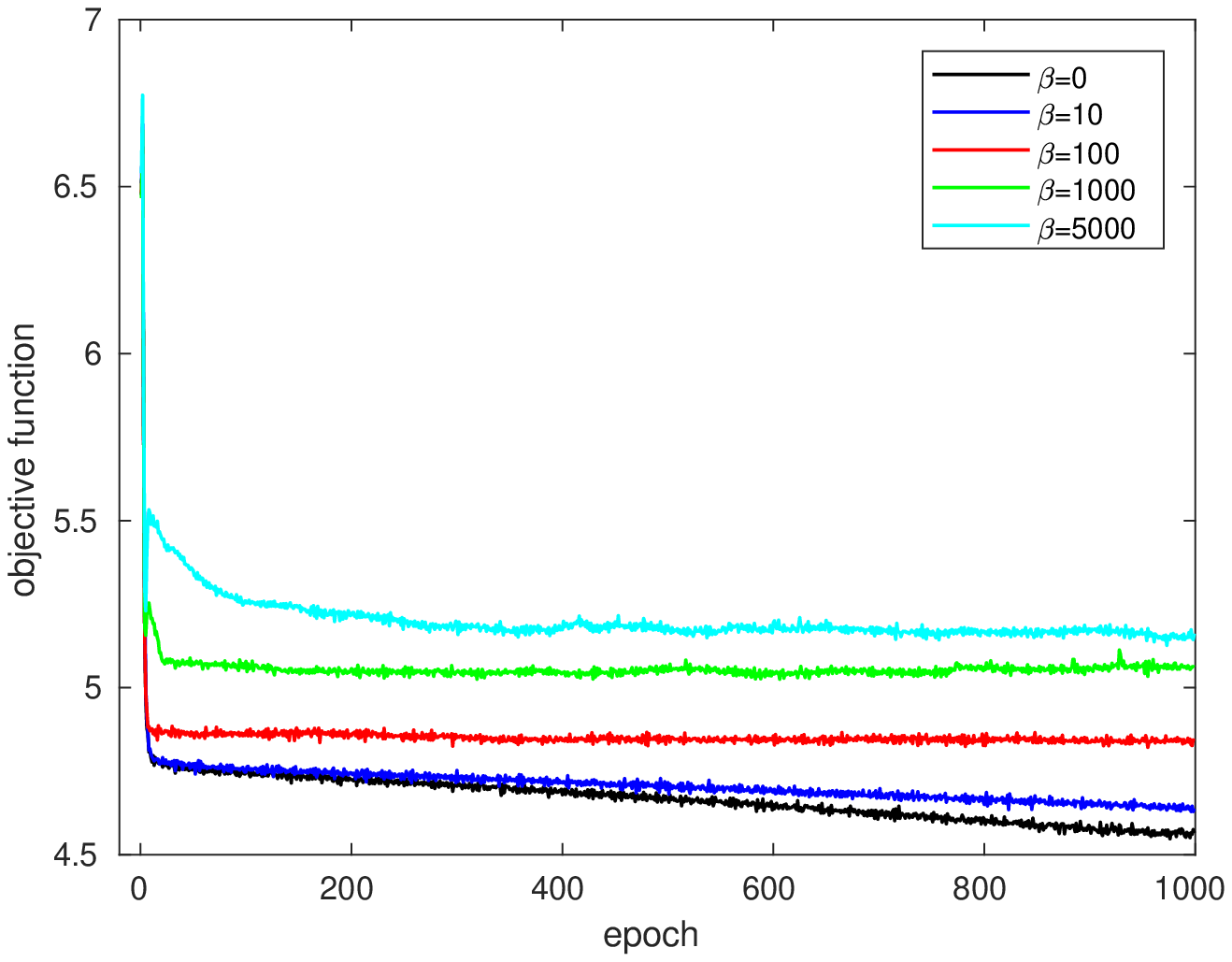}
		\includegraphics[width=0.49\textwidth]{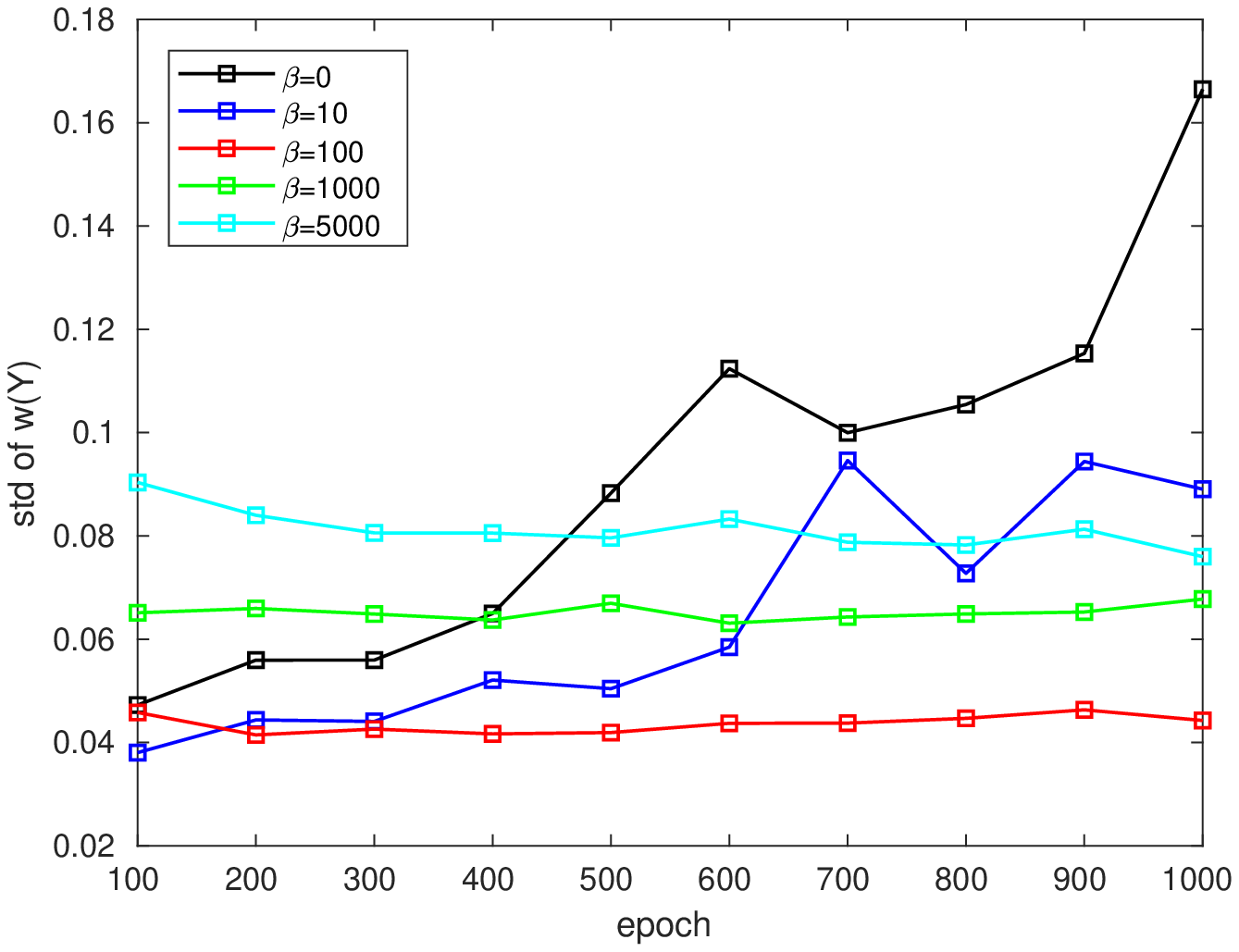}
	}
	\caption{$N=4$. Left: The evolution behavior of stochastic optimization, where the penalty term varies in terms of $\beta$. Only the cross entropy has been plotted. Right: The standard deviation of $w(\BY)$.}\label{fig:d4_v1_cong_std} 
\end{figure}

\begin{figure}
	\center{
		\includegraphics[width=0.49\textwidth]{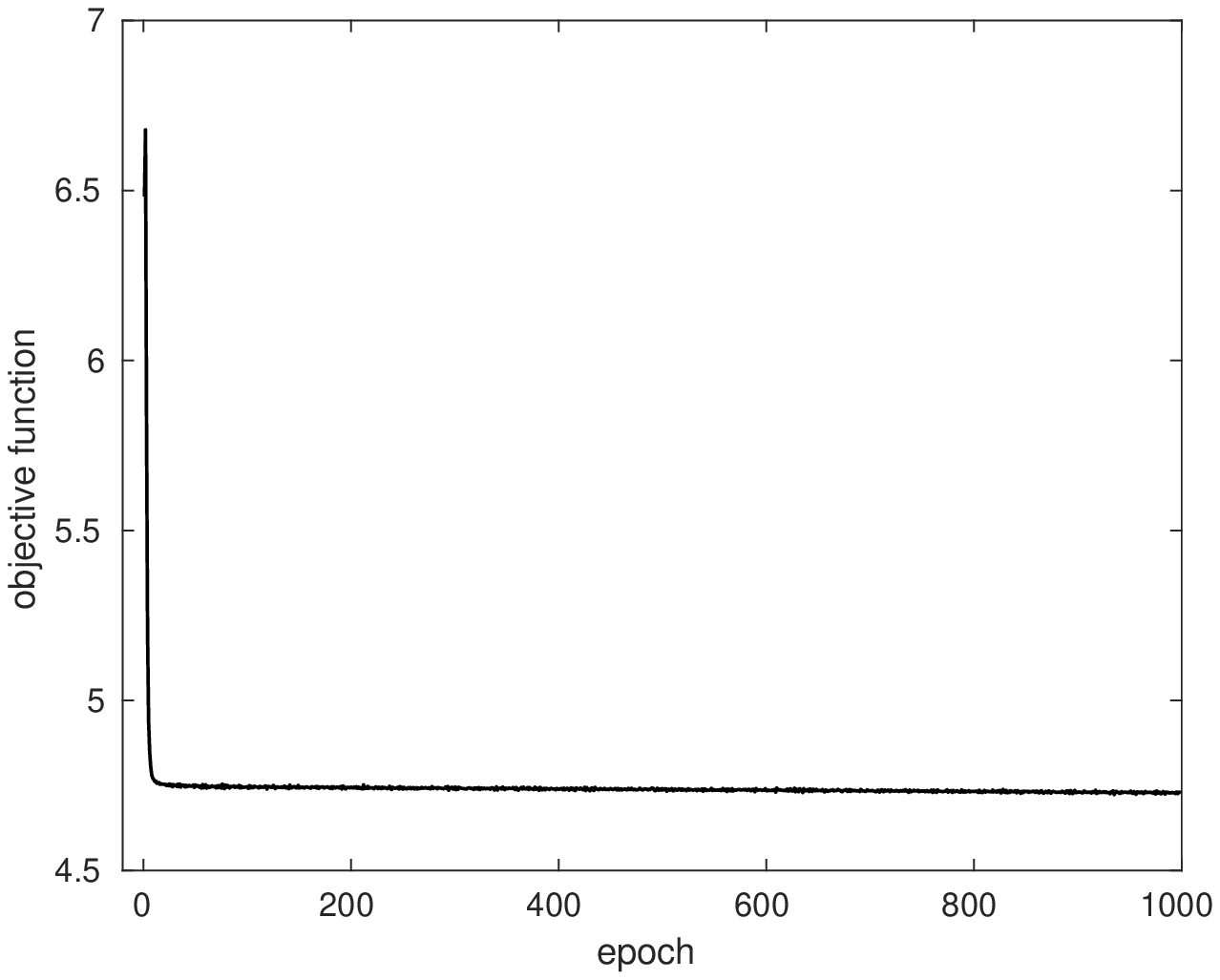}
		\includegraphics[width=0.49\textwidth]{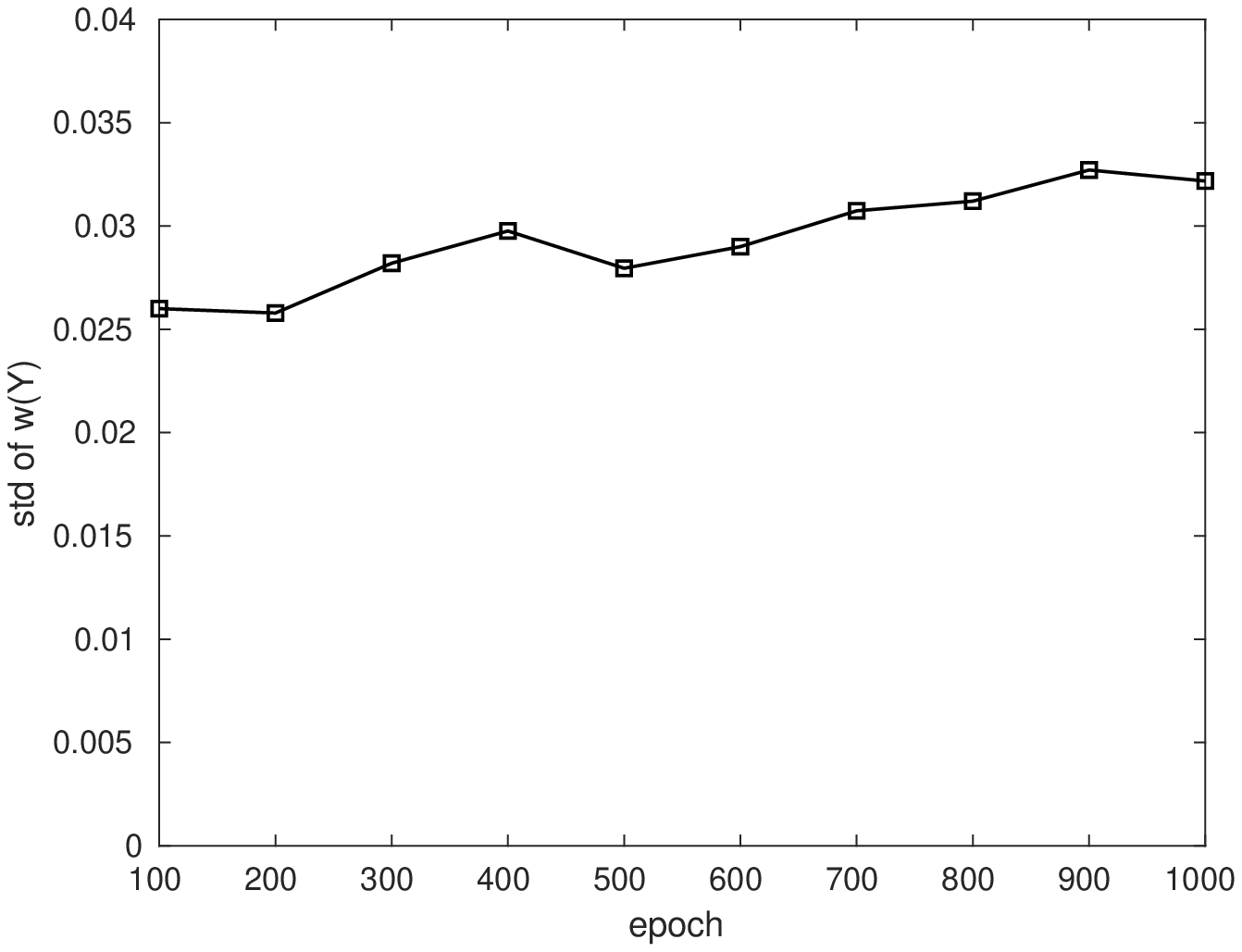}
	}
	\caption{$M=4$. Left: The evolution behavior of stochastic optimization, where the penalty term is not included in the objective function. Right: The standard deviation of $w(\BY)$.}\label{fig:d4_v2_cong_std} 
\end{figure}

We then consider a four-dimensional case, i.e., $M=4$, where $\mathbb{E}[I_B]\approx\textrm{0.121}$ and $\sigma_{I_B}\approx\textrm{0.326}$. The training set from the reduced-order model includes 12,032 samples satisfying $g(u_{M,h,c})\geq 0$, and 7,519 samples satisfying $g(u_{M,h,c})< 0$, among which only 593 samples are really missed by the {reduced-order} model, i.e., $g(u_{M,h,c})< 0$ while $g(u_{M,h,f})\geq 0$. The simulation results are given in figure \ref{fig:d4_v1_cong_std}. There are several interesting observations: First, if no penalty term is included in the objective function, the evolution of stochastic optimization has two types of behavior. The function value plummets at the beginning and then decays very slowly. This is because the size of the data set is relatively small in terms of the dimension $M$ such that the overfitting occurs. Note that for this case, the standard deviation of $\sigma_w$ increases with respect to the epoch, meaning that the efficiency of the importance sampling estimator decreases if the training of the generative model is stopped at a larger epoch. Second, when more and more penalty is included, the slow decay in the optimization iteration disappears, implying that the regularization works. Furthermore, $\sigma_w$ stops increasing after the regularization is introduced. It appears that $\sigma_w$ increases with respect to $\beta$, meaning too much regularization will deteriorate the efficiency of importance sampling estimator. Third, note that when the epoch is 100, the generative models subject to $\beta$=0 and $100$ give a comparable $\sigma_w$. This implies that early stopping may be used. However, it seems that the penalty term yields much more robustness. For $\beta=100$, $\sigma\approx0.042$, which yields that
\[
\frac{N_{\mathsf{IS}}}{N_{\mathsf{MC}}}\approx\left(\frac{0.042}{0.326}\right)^2\approx1.66\%.
\]
The other way to alleviate the overfitting is to enlarge the training set. In figure \ref{fig:d4_v2_cong_std}, we plot the results subject to a larger training set, which has 120,137 samples satisfying $g(u_{M,h,c})\geq 0$ and 81,839 samples satisfying $g(u_{M,h,c})\geq 0$. For this case, a smaller $\sigma_w$ is achieved without using any penalty term in the objective function. 

\begin{figure}
	\center{
		\includegraphics[width=0.49\textwidth]{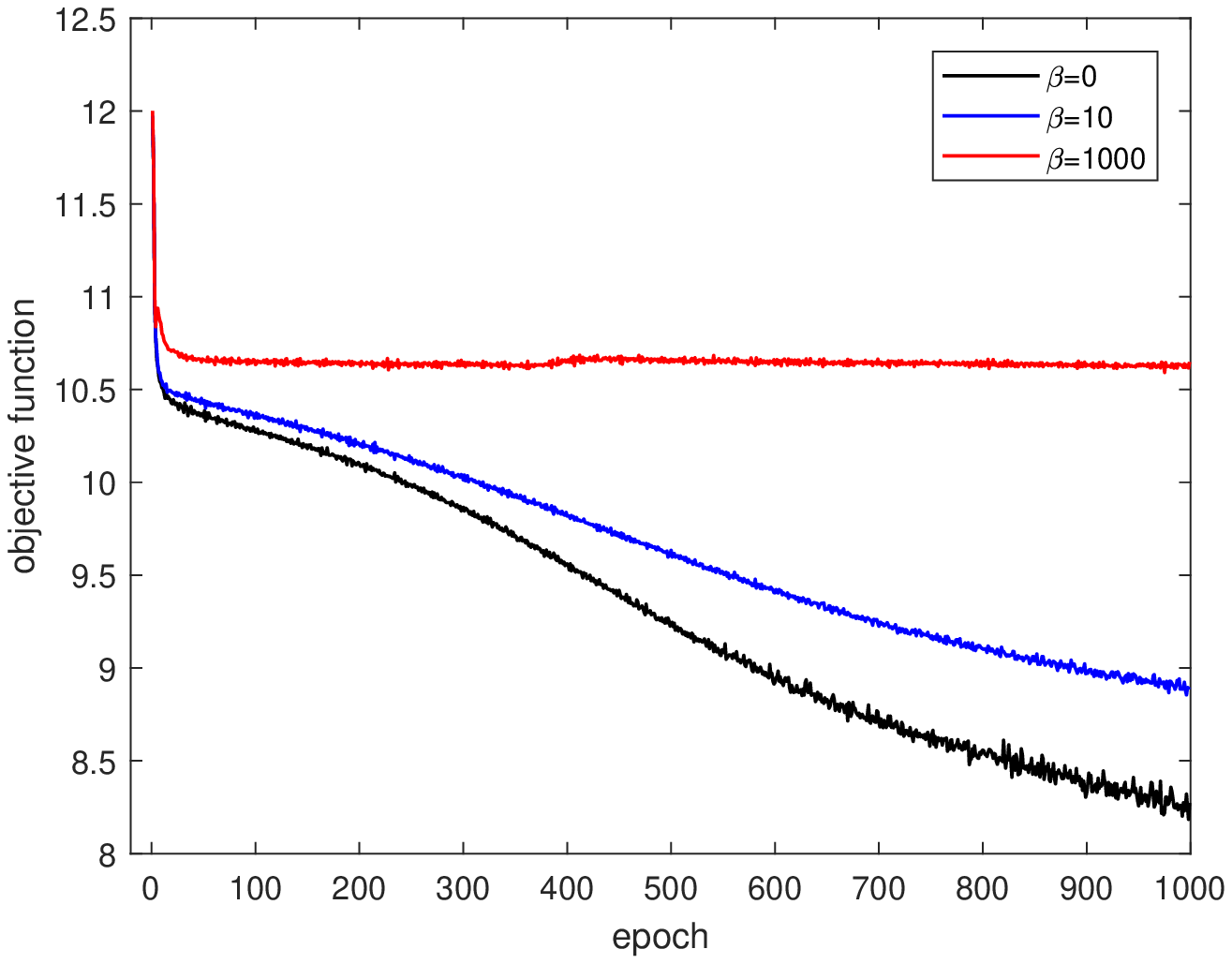}
		\includegraphics[width=0.49\textwidth]{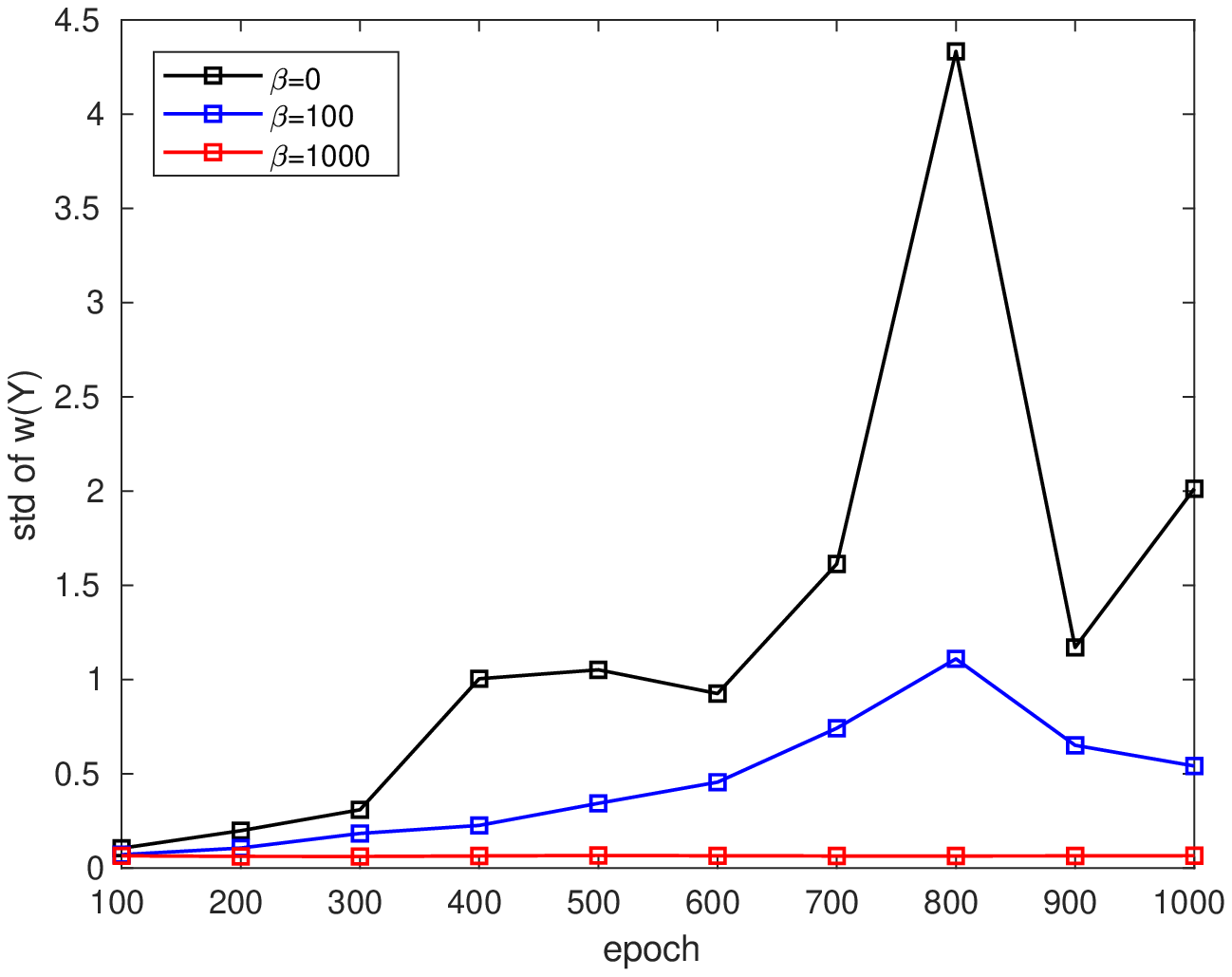}
	}
	\caption{$M=8$. Left: The evolution e behavior of stochastic optimization, where the penalty term varies in terms of $\beta$. Only the cross entropy has been plotted. Right: The standard deviation of $w(\BY)$.}\label{fig:d8_v1_cong_std} 
\end{figure}

We now look at a eight-dimensional case, i.e., $M=8$, where  $\mathbb{E}[I_B]\approx\textrm{0.130}$ and $\sigma_{I_B}\approx\textrm{0.336}$. The training set from the reduced-order model includes 12,504 samples satisfying $g(u_{M,h,c})\geq 0$, and 22,049 samples satisfying $g(u_{M,h,c})< 0$, among which only 828 samples are really missed by the reduced-order model, i.e., $g(u_{M,h,c})< 0$ while $g(u_{M,h,f})\geq 0$. The results are given in figure \ref{fig:d8_v1_cong_std}. Compared to the previous case, similar results have been observed. Since the dimension is doubled but the size of training set remains the same, it is seen that the performance of the generative model deteriorates quickly as the epoch increases if no penalty term is used. Actually, after epoch 300   $\sigma_w$ is larger than 0.33 when $\beta=0$, meaning that the importance sampling estimator is less efficient than the Monte Carlo estimator. Again, the penalty term can stabilize $\sigma_w$, which is about 0.063 for $\beta=1000$. For this case, 
\[
\frac{N_{\mathsf{IS}}}{N_{\mathsf{MC}}}\approx\left(\frac{0.063}{0.336}\right)^2\approx3.52\%.
\]  

\begin{figure}
	\center{
		\includegraphics[width=0.49\textwidth]{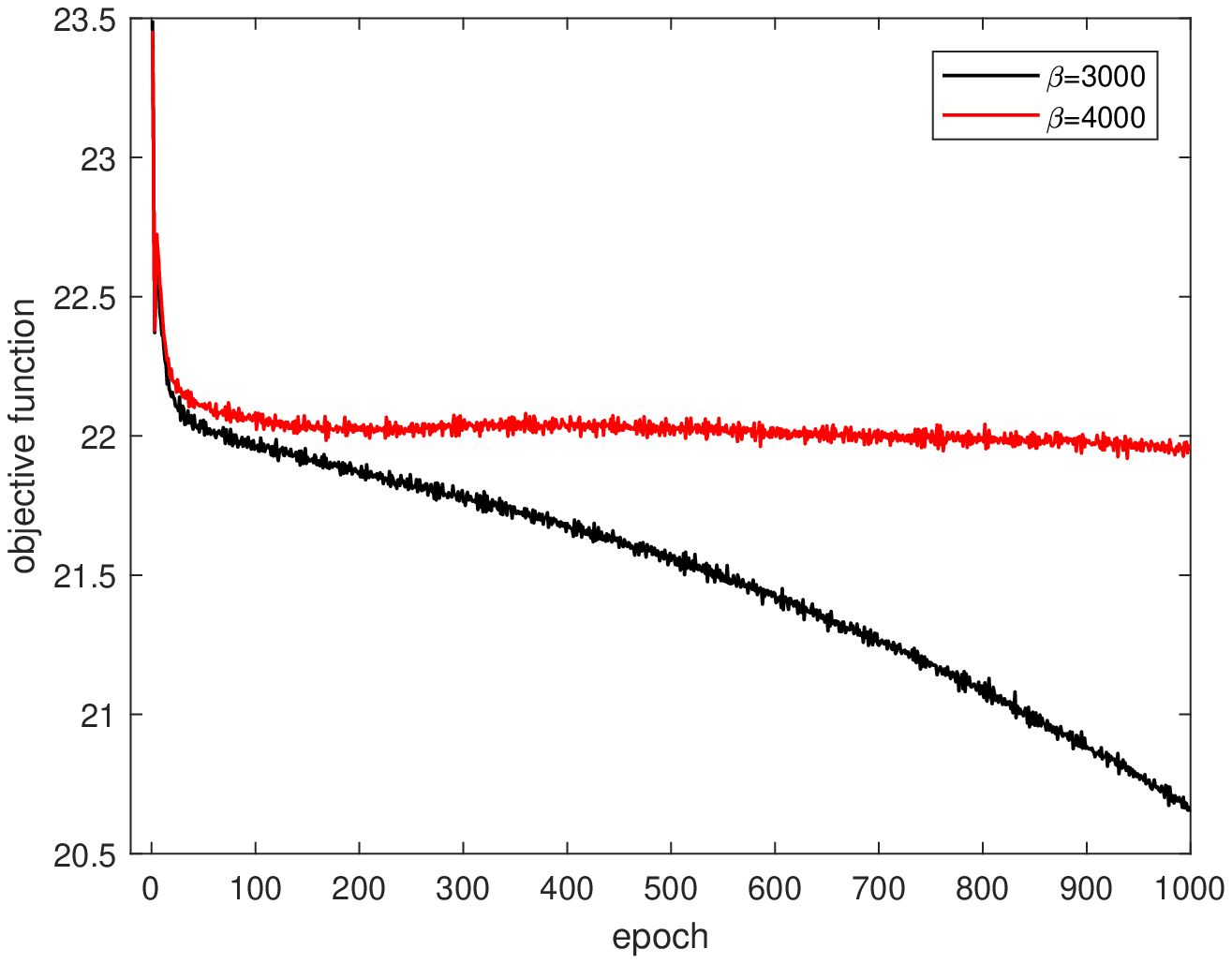}
		\includegraphics[width=0.49\textwidth]{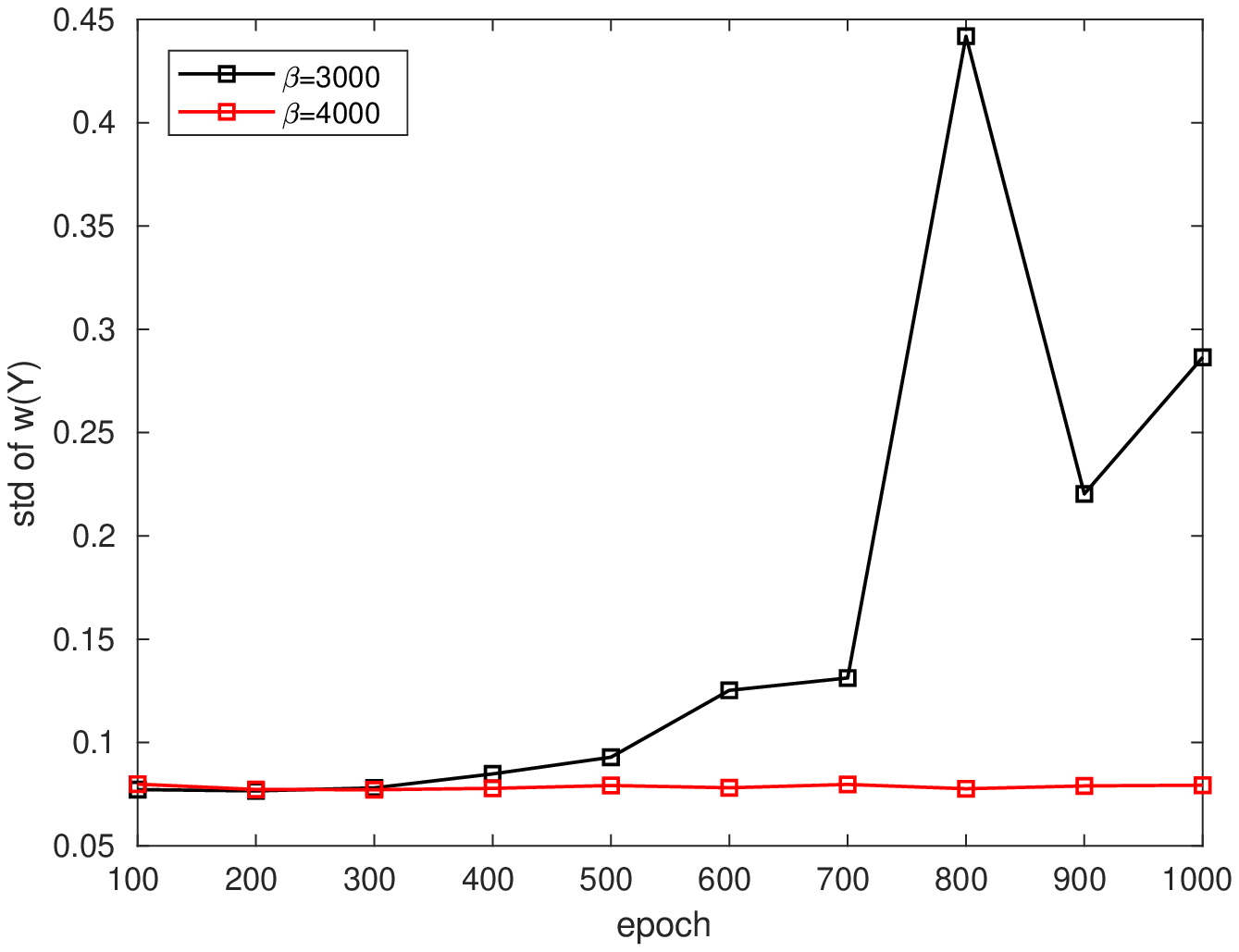}
	}
	\caption{$M=16$. Left: The evolution behavior of stochastic optimization, where the penalty term varias in terms of $\beta$. Only the cross entropy has been plotted.  Right: The standard deviation of $w(\BY)$.}\label{fig:d16_cong_std} 
\end{figure}

We double the dimension to consider $M=16$, where $\mathbb{E}[I_B]\approx0.134$ and $\sigma_w\approx0.340$. 
The training data set includes 12,975 samples that $g(u_{M,h,c})\geq 0$ and 34,847 samples that 
$g(u_{M,h,c})< 0$, among which only 913 samples are really missed by the reduced-order model. The results are plotted in figure \ref{fig:d16_cong_std}. For $\beta=4000$, we obtain $\sigma_w\approx0.078$, which yields that 
\[
\frac{N_{\mathsf{IS}}}{N_{\mathsf{MC}}}\approx\left(\frac{0.078}{0.340}\right)^2\approx5.26\%.
\]

\begin{figure}
	\center{
		\includegraphics[width=0.49\textwidth]{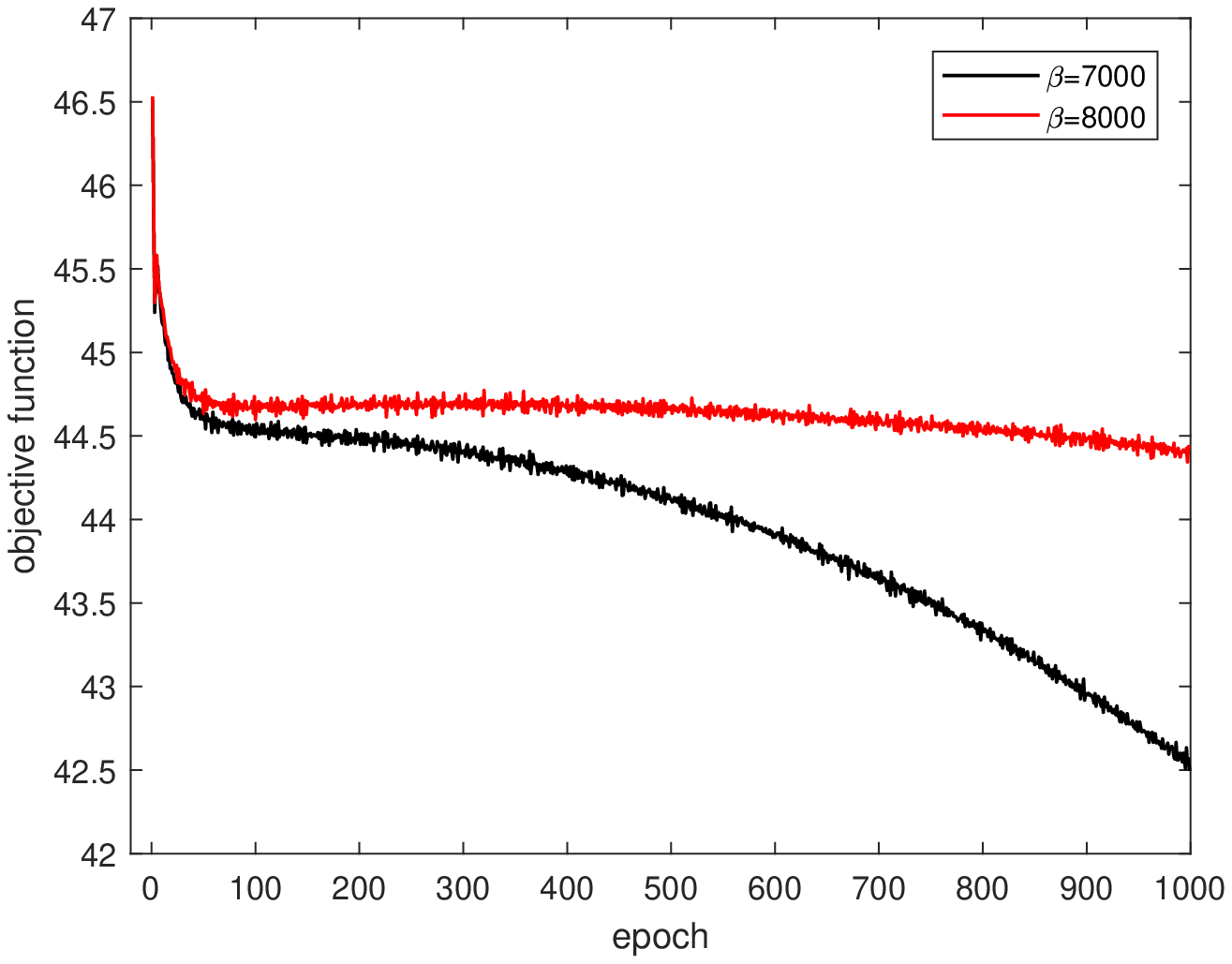}
		\includegraphics[width=0.49\textwidth]{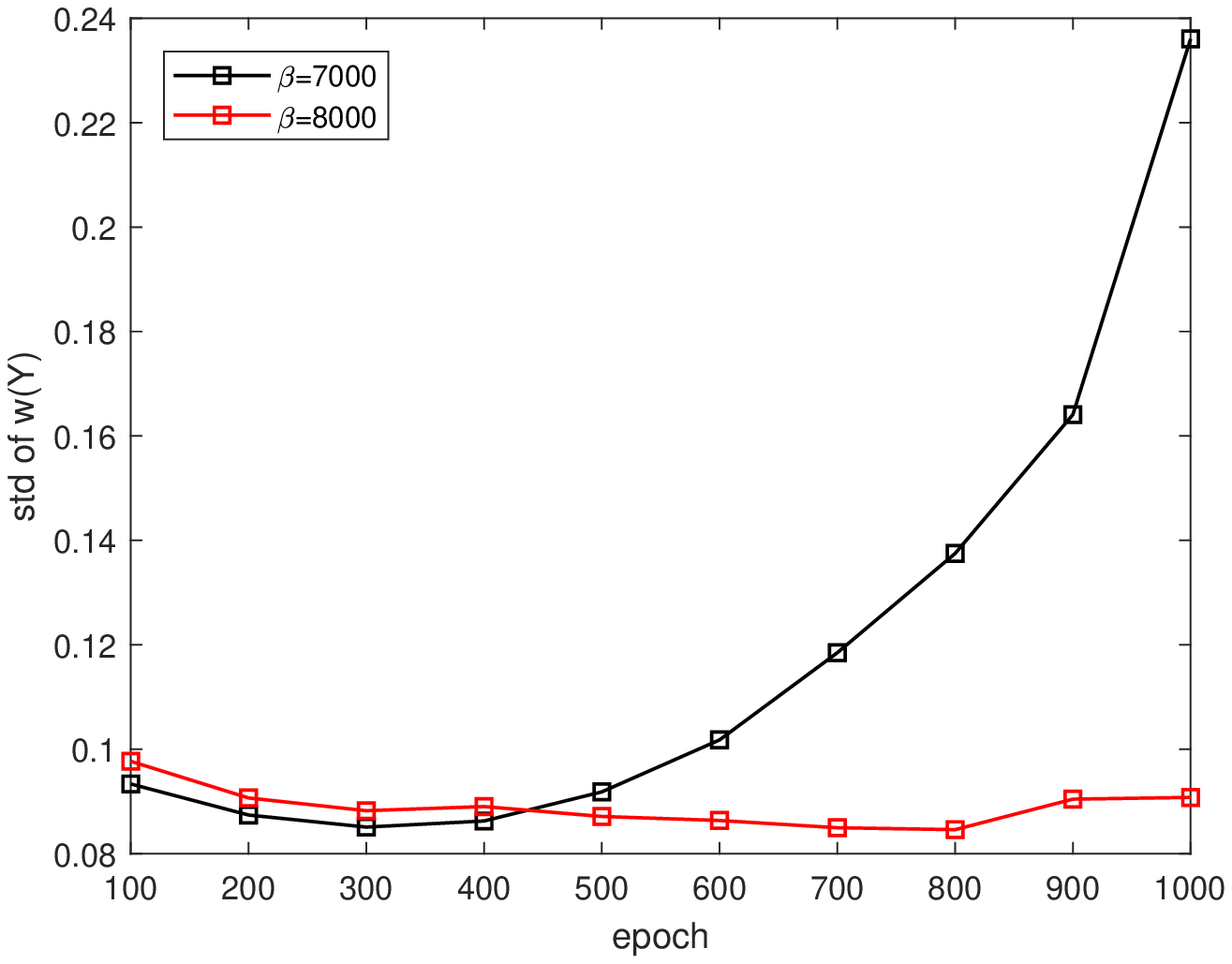}
	}
	\caption{$M=32$. Left: The evolution behavior of stochastic optimization, where the penalty term varias in terms of $\beta$. Only the cross entropy has been plotted.  Right: The standard deviation of $w(\BY)$.}\label{fig:d32_cong_std} 
\end{figure}

We finally consider a case that $M=32$, where $\mathbb{E}[I_B]\approx0.134$ and $\sigma_w\approx0.341$. The training data set includes 13,267 samples that $g(u_{M,h,c})\geq 0$ and 30,402 samples that 
$g(u_{M,h,c})< 0$, among which only 1,067 samples are really missed by the reduced-order model. Although the dimension is high and the data set is relatively small, we obtain $\sigma_w\approx0.089$ with $\beta=8000$ (see figure \ref{fig:d32_cong_std}), which yields that
\[
\frac{N_{\mathsf{IS}}}{N_{\mathsf{MC}}}\approx\left(\frac{0.089}{0.341}\right)^2\approx6.81\%.
\]

\begin{figure}
	\center{
		\includegraphics[width=0.49\textwidth]{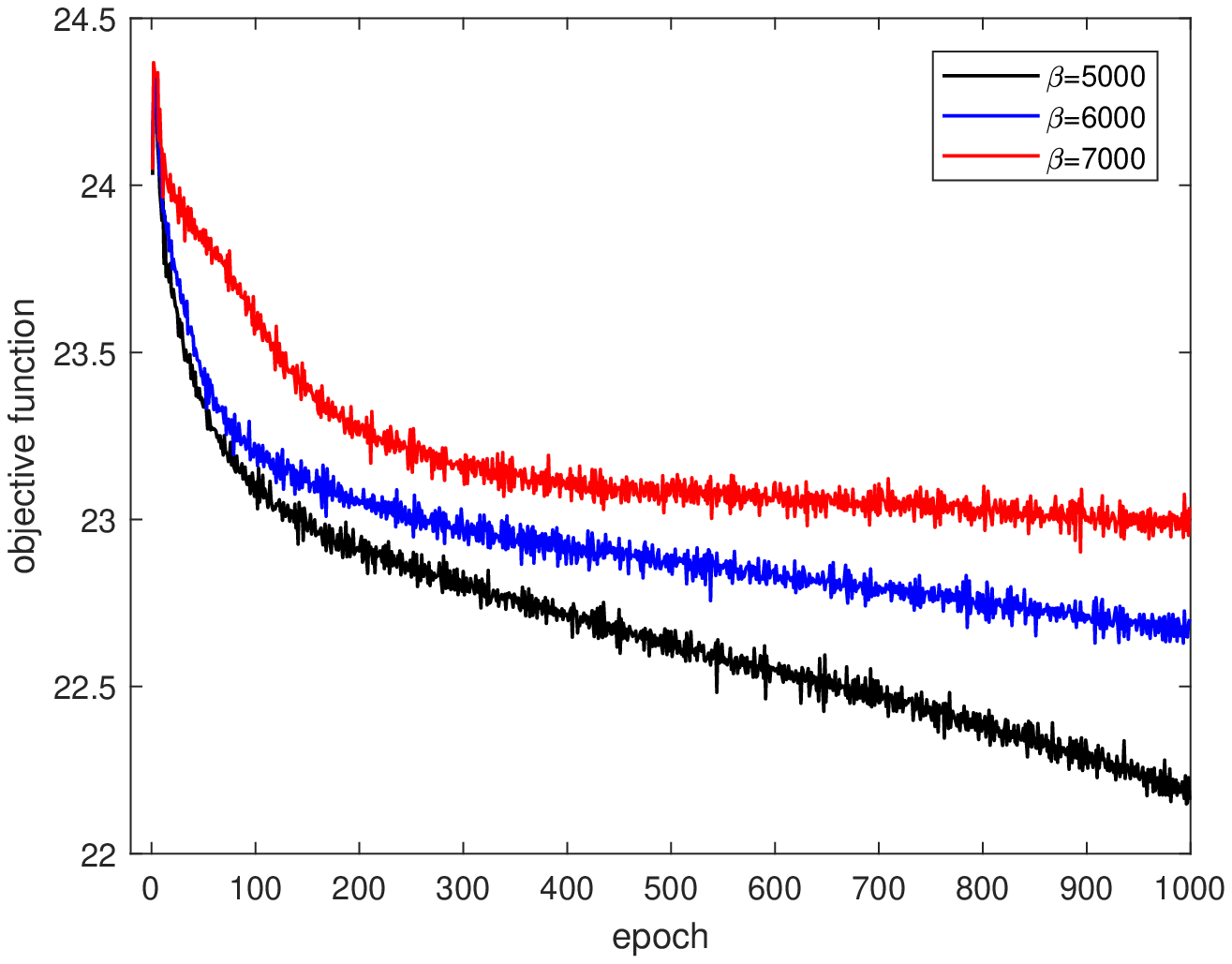}
		\includegraphics[width=0.49\textwidth]{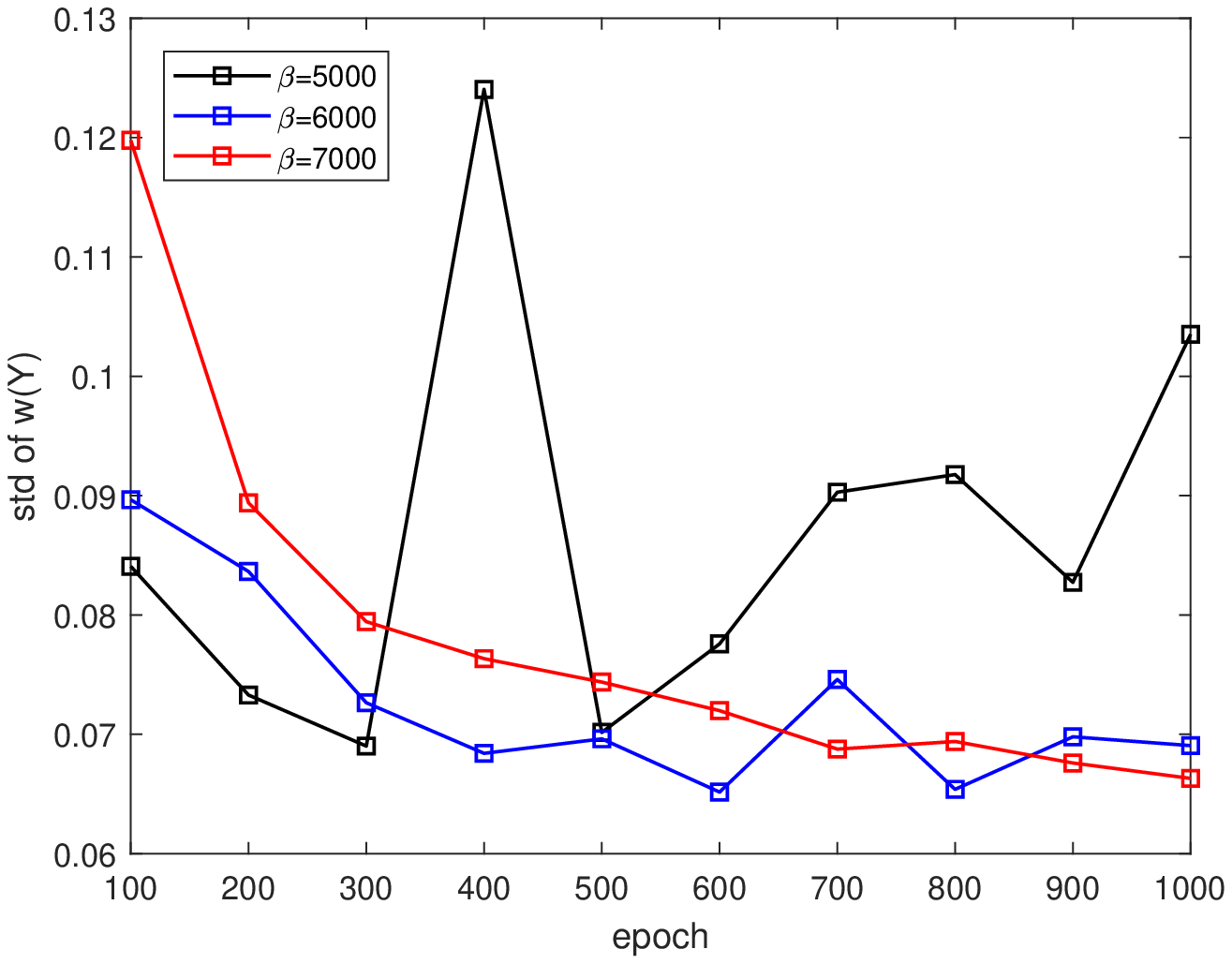}
	}
	\caption{$M=16$. Left: The evolution behavior of stochastic optimization, where the penalty term varias in terms of $\beta$. Only the cross entropy has been plotted.  Right: The standard deviation of $w(\BY)$.}\label{fig:IS2_d16_cong_std} 
\end{figure}

\begin{figure}
	\center{
		\includegraphics[width=0.49\textwidth]{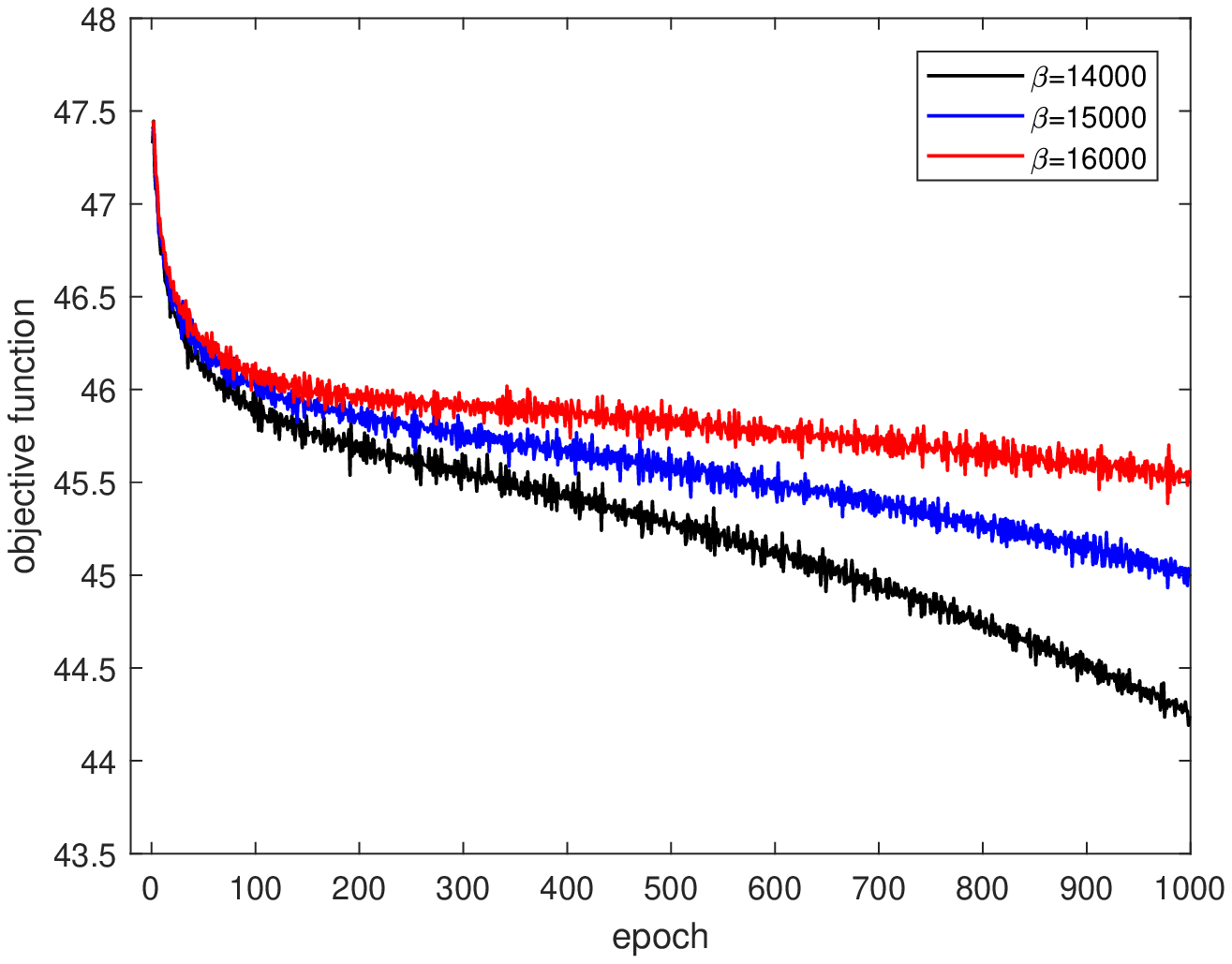}
		\includegraphics[width=0.49\textwidth]{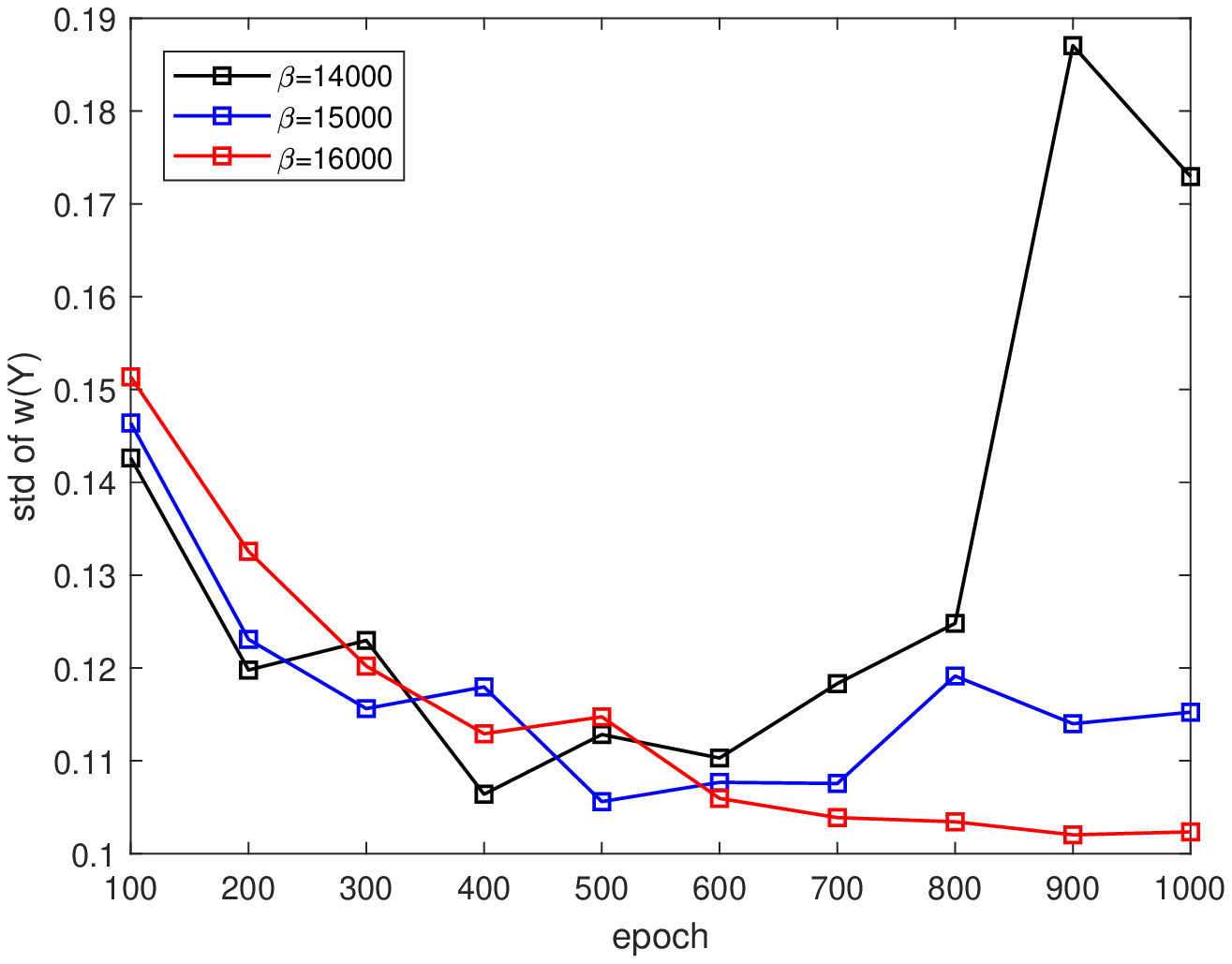}
	}
	\caption{$M=32$. Left: The evolution behavior of stochastic optimization, where the penalty term varias in terms of $\beta$. Only the cross entropy has been plotted.  Right: The standard deviation of $w(\BY)$.}\label{fig:IS3_d32_cong_std} 
\end{figure}

To this end, we have studied the performance of the generative-model-based importance sampling estimator for different random dimension $M$, where the configuration of the generative model is fixed, the ADAM method is subject to a fixed learning rate, and the size of training set remains comparable for all $M$. Very encouraging results have been obtained in terms of the ratio $N_{\mathsf{IS}}/N_{\mathsf{MC}}$ for $M$ varying from 2 to 32. 
The penalty term in the objective function appears important for the robustness of the algorithm. Note that all cases we have studied so far are subject to a relative large correlation length $l_c=1$, e.g., $\frac{\lambda_{16}}{\lambda_1}=1.22$e-3. The fast decay of the eigenvalues may reduce the difficulty of density estimation in terms of the dimensionality. To clarify this concern, we study a relatively small correlation length $l_c=0.1$ and let $M=16$, where $\frac{\lambda_{16}}{\lambda_1}=4.53\%$. Due to the slower decay of eigenvalues, the high-order modes in the Karhunen-Lo\'{e}ve expansion will play a much more role for the value {of} $\|u\|_{H^1(D)}$. We then refine the coarse mesh from 10 equidistant linear finite elements to 30. We have $\mathbb{E}[I_B]\approx0.093$ and $\sigma_w\approx0.290$. The training data set includes 9,010 samples that $g(u_{M,h,c})\geq 0$ and 40,568 samples that $g(u_{M,h,c})< 0$, among which only 609 samples are really missed by the reduced-order model. The results have been plotted in figure \ref{fig:IS2_d16_cong_std}. Compared to the previous cases with a large correlation length, the relaxation time of stochastic optimization  increases in the sense that the optimal generative model will be achieved at a larger epoch. Other than that, the results are qualitatively similar to previous observations. In particular, the penalty term is critical for robustness.  For $\beta=7000$, we are able to obtain 
\[
\frac{N_{\mathsf{IS}}}{N_{\mathsf{MC}}}\approx\left(\frac{0.071}{0.290}\right)^2\approx6.00\%.
\]
{We now let $M=32$, where $\frac{\lambda_{32}}{\lambda_1}=1.11\%$, $\mathbb{E}[I_B]\approx0.115$ and $\sigma_w\approx0.319$. The data from the reduced-order model include 11,260 samples that that $g(u_{M,h,c})\geq 0$ and 88,342 samples that $g(u_{M,h,c})< 0$, among which only 905 samples are really missed by the reduced-order model. 
It is seen that the redundant data is about eight times as many as the data that $g(u_{M,h,c})\geq 0$. This is because the high-order engenfunctions $\theta_i$ are highly oscillating, and become more important in the evaluation of $\|u\|_{H^1(D)}$ when eigenvalues decay slowly. The coarse mesh cannot capture the high oscillation well, and introduce a large error when the random variables associated with the high-order eigenfunctions take a large value. We here simply truncate the data set with respect to the value of $|g_{h,c}(\By^{(i)})|$. We only keep half of the data that $g(u_{M,h,c})< 0$, which have a smaller $|g_{h,c}(\By^{(i)})|$. Since the dependence on the high-order eigenfunctions is stronger, we increase the depth $L$ from $16$ to $24$. Other than that, all other set-up remains the same. The results are plotted in figure \ref{fig:IS3_d32_cong_std}. When the epoch is 1000, we  obtain
\[
\frac{N_{\mathsf{IS}}}{N_{\mathsf{MC}}}\approx\left(\frac{0.102}{0.319}\right)^2\approx10.22\%,
\]
with $\beta=16000$. }

\section{Summary and discussions}
In this work we have proposed a methodology to couple the reduced-order model and the generative model to construct an importance sampling estimator. Our numerical experiments show that this idea is actually feasible although the approximation of high-dimensional PDF is difficult due to the curse of dimensionality. From the application point of view, the generative models haven been trained to approximate the data distribution given by high-resolution images, where the criterion for effectiveness is quite ad hoc although the dimensionality is really high. We adapt the generative model to deal with a physical problem and measure its effectiveness rigorously through the variance reduction it is able to introduce. It appears that the generative model does have the ability to encode the information in the high-dimensional data from a physical model. However, it seems that the properties of the problem should be incorporated into the training process to enhance the robustness. For our problem, the {regularization} induced by the penalty term is much more robust than a general regularization technique in {machine} learning such as early {stopping}. We have demonstrated that the generative-model-based important sampling estimator can achieve a significant variance reduction for {at least} random dimensions of $\mathit{O}(10)$ with respect to a UQ problem. {To test the robustness, we have fixed the configuration of the generative model and the parameters of the optimization algorithm. For the problems studied, at least about $90\%$ reduction in variance is achieved for the dimension $M$ up to 32 with about $10^4$ samples.} At this moment, it is unclear how many random dimensions the generative model can effectively deal with for UQ problems in terms of the variance reduction of importance sampling. However, the scalability of deep nets makes it very promising to apply our methodology to a larger random dimension by using a larger depth $L$.  

There are many possibilities to improve the current work. For example, in all our numerical experiments, a fixed partition of the random vector is used. A more effective partition strategy can be employed especially when the number of effective random dimensions is much smaller than the total number of random dimensions.  Other generative models can also be employed. The invertible mapping has been recently introduced into a general adversarial network such that GAN is able to perform exact likelihood evaluation \cite{Grover_2018}.  In \cite{Zhang_2018}, a new flow-based generative model is proposed by incorporating the optimal transport theory.  
How these flow-based models help importance sampling in our problem setting is an interesting question. {Another possibility is to take into account the dimension reduction in the probability space such that we can mainly focus on the effective random dimensions.}

\section*{Acknowledgment}
The first author's work was supported by NSF grant DMS-1620026 and AFOSR grant FA9550-15-1-0051, and the second author's work was supported by NSF grants 1320351 and 1642991.


\begin{thebibliography}{100}
	\bibitem{Dinh_2014}
	L.~Dinh, D.~Krueger, and S.~Bengio, 
	\emph{Nice: non-linear independent components estimation}, (2014), arXiv:1410.8516.
	
	\bibitem{Dinh_2016}
	L.~Dinh, J.~Sohl-Dickstein, and S.~Bengio, 
	\emph{Density estimation using real NVP}, (2017), 
	arXiv:1605.08803v3.

	\bibitem{Giles_AN15}
	M.~B.~Giles, 
	\emph{Multilievel Monte Carlo methods}, 
	Acta Numerica, (2015), pp. 259--328.

	\bibitem{Goodfellow_2014}
	I.~Goodfellow, J.~Pouget-Abadie, M.~Mirza, B.~Xu, D.~Warde-Farley, S.~Ozair, A.~Courville, and Y.~Bengio, 
	\emph{Generative adversarial nets}. Advances in Neural Information Processing Systems, (2014), 2672--2680.
	
	\bibitem{Graves_2013}
	A.~Graves, 
	\emph{Generating sequences with recurrent neural networks}, (2013), arXiv:1308.0850.
	
    \bibitem{Grover_2018}
    A.~Grover, M.~Dhar, and S.~Ermon, 
    \emph{Flow-GAN: Combining maximum likelihood and adversarial learning in generative models}, (2018),
    arXiv:1705.08868v2.	
	
	\bibitem{Szegedy_2015}
	S.~Ioffe, and C.~Szegedy, 
	\emph{Batch normalization: Accelerating deep network training by reducing internal covariance shift}, 
	(2015), arXiv:1502.03167v3.	
	
    \bibitem{Karnia_JSC02}
    {M.~Jardak, C.-H.~Su, and G.~Karniadakis}, 
    \emph{Spectral polynomial chaos solutions of the stochastic advection equation}, 
    J. Sci. Comput., 17 (2002), pp.~319--338.	
	
	\bibitem{Dhariwal_2018}
	D.~P.~Kingma, and P.~Dhariwal, 
	\emph{Glow: Generative flow with invertable 1x1 convolutions}, (2018), arXiv:1807.03039v2.
	
	\bibitem{ADAM_2017}
	D.~P.~Kingma, and J.~L.~Ba, 
	\emph{ADAM: A method for stochastic optimization},(2017), arXiv:1412.6980v9.
	
    \bibitem{Kingma_2016}
    D.~P.~Kingma, T.~Salimans, R.~Jozefowicz, X.~Chen, I.~Sutskever, and M.~Welling, 
    \emph{Improving variational inference with inverse autoregressive flow},  Advances in Neural Information Processing Systems, (2016), pp. 4743--4751.	
	
	\bibitem{Oord_2016a}
	A.~van~den~Oord, N.~Kalchbrenner, and K.~Kavukcuoglu,
	\emph{Pixel recurrent neural networks},  (2016),
	arXiv:1601.06759.
	
	\bibitem{Oord_2016b}
	A.~van~den~Oord, N.~Kalchbrenner, O.~Vinyals, L.~Espeholt, A.~Graves, and K.~Kavukcuoglu,
	\emph{Conditional image generation with PixcelCNN decoders},  (2016),
	arXiv:1606.05328.
	
	\bibitem{Papamakarios_2018}
	G.~Papamakarios, T.~Pavlakou, and I.~Murray, 
	\emph{Masked autoregressive flow for density estimation}, (2018), 
	arXiv:1705.07057v4.
	
	\bibitem{Scott2015}
	D.~Scott, 
	Multivariate Density Estimation: Theory, Practice, and Visualization, 
	2nd Edition, John Wiley \& Sons, Inc., 2015.
	
    \bibitem{Wan_model3}
    {X.~Wan, and B.~L.~Rozovskii}, 
    \emph{The Wick-Malliavin approximation of elliptic problems with log-normal 
    random coefficients}, SIAM J. Sci. Comput., 35(5) (2013), pp.~A2370--A2392.
	
	
	\bibitem{Zhang_2018}
	L.~Zhang, W.~E, and L.~Wang, 
	\emph{Monge-\mbox{A}mp\'{e}re flow for generative modeling}, (2018), 
	arXiv:1809.10188v1.
	
	\bibitem{Zhang_AS05}
	T.~Zhang and B.~Yu,
	\emph{Boosting with early stopping: convergence and consistency}, 
	Ann. Statist., 33(4) (2005), pp. 1538--1579.

\end{thebibliography}
\end{document}